\documentclass[acmlarge]{acmart}

\usepackage{multirow}
\usepackage[figuresright]{rotating}
\usepackage{graphicx}
\usepackage{colortbl}
\usepackage{pifont}
\usepackage{threeparttable}

\AtBeginDocument{%
  }

\setcopyright{acmlicensed}
\copyrightyear{2018}
\acmYear{2018}
\acmDOI{XXXXXXX.XXXXXXX}

\acmJournal{POMACS}
\acmVolume{37}
\acmNumber{4}
\acmArticle{111}
\acmMonth{8}

\begin{document}

%%
%% The ''title'' command has an optional parameter,
%% allowing the author to define a ''short title'' to be used in page headers.
\title{New Emerged Security and Privacy of Pre-trained Model: a Survey and Outlook}

%%
%% The ''author'' command and its associated commands are used to define
%% ''authornote'' and ''authornotemark'' commands

\author{Meng Yang}
\affiliation{%
  \institution{University of Technology Sydney}
  \country{Australia}}
\email{meng.yang-4@student.uts.edu.au}

\author{Tianqing Zhu}
\affiliation{%
  \institution{City University of Macao}
  \country{China}}
\email{tqzhu@cityu.edu.mo}

\author{Chi Liu}
\affiliation{%
  \institution{City University of Macao}
  \country{China}}
\email{chiliu@cityu.edu.mo}

\author{Wanlei Zhou}
\affiliation{%
  \institution{City University of Macao}
  \country{China}}
\email{wlzhou@cityu.edu.mo}

\author{Shui Yu}
\affiliation{%
  \institution{University of Technology Sydney}
  \country{Australia}}
\email{Shui.Yu@uts.edu.au}

\author{Philip S. Yu}
\affiliation{%
  \institution{University of Illinois at Chicago}
  \country{United States}}
\email{psyu@uic.edu}

\renewcommand{\shortauthors}{Meng et al.}

\begin{abstract}
  Thanks to the explosive growth of data and the development of computational resources, it is possible to build pre-trained models that can achieve outstanding performance on various tasks, such as neural language processing, computer vision, and more. 
  Despite their powerful capabilities, pre-trained models have also sparked attention to the emerging security challenges associated with their real-world applications.
  Security and privacy issues, such as leaking privacy information and generating harmful responses, have seriously undermined users' confidence in these powerful models.
  Concerns are growing as model performance improves dramatically.
  Researchers are eager to explore the unique security and privacy issues that have emerged, their distinguishing factors, and how to defend against them. However, the current literature lacks a clear taxonomy of emerging attacks and defenses for pre-trained models, which hinders a high-level and comprehensive understanding of these questions.
  To fill the gap, we conduct a systematical survey on the security risks of pre-trained models, proposing a taxonomy of attack and defense methods based on the accessibility of pre-trained models' input and weights in various security test scenarios. This taxonomy categorizes attacks and defenses into No-Change, Input-Change, and Model-Change approaches. With the taxonomy analysis, we capture the unique security and privacy issues of pre-trained models, categorizing and summarizing existing security issues based on their characteristics. In addition, we offer a timely and comprehensive review of each category's strengths and limitations. Our survey concludes by highlighting potential new research opportunities in the security and privacy of pre-trained models. 
\end{abstract}




\received{20 February 2007}
\received[revised]{12 March 2009}
\received[accepted]{5 June 2009}

\maketitle

\section{Introduction}
The recent years have seen rapid advancements in large artificial intelligence (AI) models, such as large language models and large vision models, leading to the concept of pre-trained models. Today, pre-trained models are viewed as large-scale encoders with sophisticated architectures and vast numbers of parameters, initially trained with advanced objectives using extensive datasets. Due to the explosive growth of data and advancements in computing technology, researchers can now scale training data and model architecture to create pre-trained models, such as GPT \cite{radford2018improvingGPT1, gpt-2} and BERT \cite{BERT2019}, that store foundational knowledge in their huge parameters. By fine-tuning these models for specific tasks, the rich knowledge they encode can enhance various downstream applications. As a result, using pre-trained models has become a common practice in developing and improving task-specific models.

While enjoying the convenience brought by pre-trained models, people are increasingly worried about potential security risks. For example, membership inference attacks \cite{75}\cite{43} can reveal information about specific contents of a training dataset, and jailbreak attacks \cite{1}\cite{3}\cite{15} can mislead model to generate harmful responses. 
However, in addition to various safety issues that have been widely studied in traditional models,  the powerful capabilities of pre-trained models also bring new safety issues that do not exist in traditional models. Thus, it is important to fill these gaps and establish higher standards for pre-trained model protection.

Some studies have been conducted to summarize the security and privacy issues in large models. However, few of them provide deep and comprehensive insight into the root causes of new safety issues in large models. 
We find that these new safety issues are introduced from the different training strategies and the large-scale dataset. 
Due to these reasons, there has been a huge gap between pre-trained models, which can be applied to different tasks, and traditional models, which focus on one specific task. For example, the ``pre-trained/fine-tune/inference'' strategy is popular in the current research area compared to the traditional ``training/inference'' strategy, wherein novel attacks emerge to attack pre-trained models in the fine-tuning process. These methods may be in line with traditional attack methods, but there are still updates in specific details, such as attacking the special training strategy of pre-trained models like Reinforcement Learning with Human Feedback (RLHF) \cite{ouyang2022training}.

There are still variations within pre-trained models. Most notably, larger models tend to exhibit stronger functionality. 
This raises the question: \textbf{what unique security and privacy issues arise as model size increases, and why do these issues occur?} Answering this is a long-term endeavor. Although it remains challenging to fully explain why large models have powerful capabilities, an initial approach could involve examining security and privacy issues in smaller pre-trained models and exploring how they generalize to larger models. This would help identify common and differing security and privacy issues, as well as similarities and differences in attack and defense strategies across model sizes.

There are various concepts and multiple attack/defense methods in this field, and the boundary between pre-trained models and traditional models is often blurred. These challenges prompted us to conduct a comprehensive survey to summarize, analyze, and classify security issues of pre-trained models, along with the corresponding defensive countermeasures.  
Our analysis points out the unique characteristics of various attacks and defenses in this domain and the differences between different methods. In addition, we propose a novel taxonomy to categorize the state-of-the-art methods in the literature. Our contributions can be itemized as follows:
\begin{itemize}
    \item We proposed a novel taxonomy of current attack/defense techniques to pre-trained models based on their attack/defense stages and specific strategies.
    \item We comprehensively summarized state-of-the-art attack/defense techniques based on the proposed taxonomy, showing their benefits and shortcomings.
    \item We reviewed methods for attacking and defending pre-trained models of different scales, summarizing their commonalities and differences.
    \item We provided critical and deep discussions on the open security and privacy issues in pre-trained models, as well as pointing out possible further research directions.
\end{itemize}

We hope this work will help to comprehensively review and evaluate the security and privacy risks of pre-trained models. 
%If a standardized evaluation system can be established, the risk assessment of pre-trained models can be more accurately performed. As a result, confidence in pre-trained models would increase for users.
Establishing a standardized evaluation system would allow for more accurate risk assessments, ultimately increasing users' confidence in pre-trained models.

\section{Preliminary}

% \begin{table}[ht]
%   \caption{Notations.}
%   \label{tab:notaion}
%   \begin{tabular}{lll}
%     \toprule
%     Notations & Symbols & Definition \\
%     \midrule
%     Target model & $E$ & original clean model from model provider  \\
%     Victim model & $E^{*}$ & victim model modified by attacker \\
%     Surrogate model  & $E_{s}$ & models built by attackers/defenders to assist their work \\
    
%     Training dataset & $D$ & dataset used to train target model \\
%     Surrogate dataset & $D_{s}$ & dataset used to attack/defense target model\\
%     Clean input & $x_{i}$, $x_{t}$ & clean image/text input sample \\ 
%     Malicious input & $x_{i}^{*}$, $x_{t}^{*}$ & image/text input sample with malicious content\\ 
%     Clean Output & $y$, $f$ & the corresponding output label/feature to clean input\\ 
%     Malicious Output & $y^{*}$, $f^{*}$ & the corresponding output label/feature to malicious input\\ 
%     Perturbation & $\Delta$ & Triggers or noises that stamped with clean input  \\
%     % Data Augmentation & $\mathcal{A}$ & Data Augmentation function work on input samples like flip, resize\\
%     % \midrule
%     % Small pre-trained Model & SPM & small pre-trained model like CLIP, BERT, GPT-1/2  \\
%     % Large pre-trained Model & large pre-trained model & large pre-trained model like GPT-3, LLaMA, Vicuna  \\
%     % Large Language Model & LLM & text-modality only large pre-trained model \\
%     % Vision-Language Model & VLM & multi-modality large pre-trained model\\
%     \bottomrule
% \end{tabular}
% \end{table}

\subsection{Definition of pre-trained model}

\subsubsection{The life process of pre-trained model}
In this survey, we divide the life process of pre-trained models into three stages, which are the pre-training stage, fine-tuning stage, and inference stage. 
% Also, we have defined and listed the notation in Table~\ref{tab:notaion}.

\begin{figure}[ht]
  \centering
  \includegraphics[width=1\linewidth]{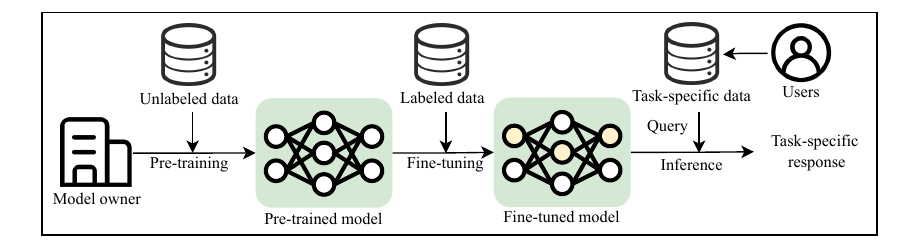}
  \caption{The life process of pre-trained models.}
  \label{fig:life_cycle_pre-trained_model}
\end{figure}

\paragraph{Pre-training stage}
In the pre-training stage, model owners build powerful models from scratch. 
To do this, they update model weights using pre-trained knowledge (such as datasets, training strategies, and parameter settings). 
As illustrated in Figure~\ref{fig:life_cycle_pre-trained_model}, model owners first collect a large dataset from diverse sources. 
The goal of the pre-training stage is to update model weights, enabling the model to learn from the collected data.
During pre-training, the model is initialized with random weight and is optimized using carefully designed loss functions. The whole training process usually lasts from weeks to months, depending on the dataset size and model size. 
After pre-training, model owners will obtain a pre-trained model that learns general knowledge well.

\paragraph{Fine-tuning stage}
Although pre-trained models have powerful performance on general knowledge, they still lack the ability to solve specific tasks, especially those tasks that are different from pre-trained tasks. 
To process a specific task, users can apply a fine-tuning strategy to update the pre-trained model. As shown in Figure~\ref{fig:life_cycle_pre-trained_model}, model owners can update entire model layers (full parameter fine-tuning) or only update some layers while keeping others frozen (partial-parameter fine-tuning) using data collected for specific purposes. Full parameter fine-tuning is usually used when the model size is small. However, as pre-trained models become larger and larger, partial-parameter fine-tuning is more practical. Current popular methods include transfer-learning \cite{TransferLearning}, LoRA \cite{hu2022lora}, and tuning strategies \cite{prefixTuning}\cite{prompttuning}.
The cost of fine-tuning a pre-trained model is less than training a new model. 

\paragraph{Inference stage}

The process of the inference stage is shown in Figure~\ref{fig:life_cycle_pre-trained_model}, model users apply the model to complete customized tasks by giving task-specific input samples without modifying the model. 
There are two scenarios: the first is when the model can directly complete a customized task, meaning the task is one of the pre-trained or downstream tasks. This typically occurs in small pre-trained models.
The second scenario requires the model user to make minor modifications to the input, allowing the pre-trained model to adapt to the customized task. This situation typically occurs with large pre-trained models capable of completing various tasks. These large models require the user to provide ``prompts'' to tell the model what task to do. For example, in an image classification task, users can design the prompt as ``What is the object in the image?'' with an input image.

\subsubsection{The development of pre-trained models}

\begin{table}
    \centering
    \caption{Pre-trained models which are regarded as target models in the security area and current state-of-the-art models which will be target models in the future. }
    \label{tab:Development_of_pretrained_model}
    \begin{tabular}{lccccc}
    \toprule
    Model & Available & Modal & Size & Base Model & Release Time \\
    \midrule
    %Resnet \cite{resnet} & open-source & image & - & - & Dec-2015 \\
    %Transformer \cite{transformer} & open-source & text & - & - & Jun-2017 \\ 
    GPT-1 \cite{radford2018improvingGPT1} & open-source & text & - & Transformer (decoder) & Jun-2018 \\
    BERT \cite{BERT2019} & open-source & text & 330M & Transformer (encoder) & Oct-2018 \\
    GPT-2 \cite{gpt-2} & open-source & text & 1.5B & GPT & Feb-2019 \\
    RoBERTa \cite{liu2019roberta} & open-source & text & 355M & BERT & Aug-2019 \\
    ALBERT \cite{lan2019albert} & open-source & text & 235M & BERT & Sep-2019 \\
    BART \cite{lewis2019bart} & open-source & text & 400M & BERT & Oct-2019 \\
    T5 \cite{raffel2020exploringT5} & open-source & text & 3B/11B & Transformer & Oct-2019 \\
    GPT-J \cite{Wanggpt-j} & open-source & text & 6B & GPT-2 & May-2021 \\
    %GPT-Neo &&&&& Mar-2021\\
    %GPT-NeoX \cite{black-etal-2022-gpt} & open-source & text & 20B & GPT-3 & Apr-2022 \\
    \midrule
    CLIP \cite{CLIPModel} & open-source & multi & 400M & ResNet/Transformer(encoder) & Jan-2021 \\
    BLIP \cite{li2022blip} & open-source & multi & 400M & Transformer & Mar-2022 \\
    OFA \cite{wang2022ofa} & open-source & multi & 33M-930M & ResNet & Jun-2022 \\
    \midrule
    \midrule
    GPT-3 \cite{gpt3} & close-source & text & 6B/175B & GPT-2 & May-2020 \\
    GPT-3.5 \cite{gpt3} & close-source & text & - & GPT-3 & Mar-2022 \\
    Instruct-GPT \cite{ouyang2022training} & close-source & text & 6B/175B & GPT-3 & Mar-2022 \\ 
    PaLM \cite{chowdhery2023palm} & close-source & text & 8B/62B/540B & Transformer (decoder) & Apr-2022 \\
    OPT \cite{zhang2022opt} & open-source & text & 125M-175B & Transformer (decoder) & May-2022 \\
    ChatGPT \cite{gpt3} & close-source & text & Application & GPT-3.5 & Nov-2022 \\
    LLaMA \cite{touvron2023llama} & open-source & text & 7B/13B/33B/65B & Transformer  (decoder) & Feb-2023 \\
    Alpaca \cite{StanfordAlpaca2023} & open-source & text & 7B/13B & LLaMA & Mar-2023\\
    Vicuna \cite{chiang2023vicuna} & open-source & text & 7B/13B & LLaMA & Mar-2023 \\
    Claude & close-source & text & Application & - & Mar-2023 \\
    Dolly \cite{DatabricksBlog2023DollyV2} & open-source &  text & 7B/12B & GPT-3 & Mar-2023 \\
    Bard (Gemini) & close-source & text & Application & PaLM & Mar-2023 \\  
    PaLM-2 \cite{anil2023palm2} & close-source & text & 16B & PaLM & May-2023 \\
    LLaMA-2 \cite{touvron2023llama2} & open-source & text & 7B/13B/70B & - & Jul-2023 \\
    Claude-2 & close-source & text & Application & - & Jul-2023 \\
    Falcon \cite{almazrouei2023falcon} & open-source & text & 7B/40B/180B & GPT-3 & Nov-2023 \\
    LLaMA-3 & open-source & text & 8B/70B & - & Apr-2024 \\
    New Bing & close-source & - & Application & GPT-4 & - \\
    \midrule

    Flamingo \cite{alayrac2022flamingo} & close-source & multi & 3B/9B/80B & CLIP+Transformer (decoder) & Apr-2022 \\
    BLIP-2 \cite{li2023blip2} & open-source & multi & 3B-12B & CLIP+OPT/T5 & Feb-2023 \\
    GPT-4 \cite{achiam2023gpt4} & close-source & multi & - & GPT-3.5 & Mar-2023 \\
    MiniGPT-4 \cite{zhu2023minigpt} & open-source & multi & 7B/13B & Vicuna/LLaMA & Apr-2023 \\
    Instruct BLIP \cite{instructblip} & open-source & multi & 7B/13B &  BLIP-2 & Jun-2023 \\
    OpenFlamingo \cite{awadalla2023openflamingo} & open-source & multi & 3B/9B & CLIP+LLaMA & Aug-2023\\
    LLaVA \cite{llavamodel} & open-source & multi & 7B/13B & Vicuna & Dec-2023 \\
    LLaVA-NeXT \cite{liu2024llavanext} & open-source & multi & 34B & LLaVA & Jan-2024\\
    GPT-4o & close-source & multi & - & GPT-4 & May-2024\\
    Claude-3 & close-source & multi & Application & - & Mar-2024 \\
    Claude-3.5 & close-source & multi & Application & - & Jun-2024 \\
    LLaMA-3.2 & open-source & multi & 11B/90B & - & Sep-2024 \\
    \bottomrule
    \end{tabular}
\end{table}

We summarize all pre-trained models which are regarded as target models in the security area and current state-of-the-art models which will be target models in the future in Table~\ref{tab:Development_of_pretrained_model}.
These models are based on ResNet \cite{resnet} or Transformer \cite{transformer}, which are two famous traditional models in the computer vision (CV) and Neutral Language Process (NLP) domain. Their architecture (encoder, decoder, skip-connection) and training strategy have hugely influenced the following work in the machine learning area. 
Based on model size and model ability, the development of pre-trained models can be divided into small pre-trained models and large pre-trained models.

\paragraph{Small pre-trained model}

Models that are small enough to be fine-tuned on consumer GPUs are defined as small pre-trained models.
We first introduce two famous language-only pre-trained models, GPT and BERT series models.
\begin{itemize}
    \item GPT-series: GPT-1 \cite{radford2018improvingGPT1} is the first GPT-series model based on Transformer-decoder, and adopted a hybrid approach of unsupervised pre-training and supervised fine-tuning. It can predict the next word of an input sentence and has set up the core architecture for GPT-series models.
    Following a similar structure, GPT-2 \cite{gpt-2} scales the model size to 1.5B. It sought to perform tasks via unsupervised language modeling without explicit fine-tuning using labeled data. 
    \item BERT-series: BERT \cite{BERT2019}, based on Transformer-encoder, focuses on capturing the context-aware word representations, which is better than the GPT model at first. This study has inspired a large number of follow-up work. For example, RoBERTa \cite{liu2019roberta} and ALBERT \cite{lan2019albert} improved the pre-training strategies, while 
    BART \cite{lewis2019bart} can be seen as generalizing BERT (due to the bidirectional encoder), GPT (with the left-to-right decoder), and many other more recent pre-training schemes. 
\end{itemize}

At the same time, researchers aim to build a vision-language model that can understand the relationship between image and text.
\begin{itemize}
    \item The most famous model is named CLIP \cite{CLIPModel}, which focuses on matching images and text using contrastive learning and is particularly strong in zero-shot learning and classification tasks.
    BLIP \cite{li2022blip} incorporates both contrastive and generative learning and is more versatile for tasks involving language generation based on images, like captioning and Visual Question Answering.
    Meanwhile, OFA \cite{wang2022ofa} is a Task-Agnostic and Modality-Agnostic framework supporting Task Comprehensiveness.
\end{itemize}

\paragraph{Large pre-trained model}

Researchers have discovered that scaling pre-trained models often leads to improved performance across various tasks \cite{zhao2023survey}. As model scale increases, their capabilities have led to unexpected advancements. We define models that can complete multiple tasks and have a significant size as large pre-trained models

\begin{itemize}
    \item GPT-series: GPT-J \cite{Wanggpt-j} aims to increase the program ability of GPT-2 by adding public software repositories into the training dataset so that the model can learn how to code a program.
    GPT-3 \cite{gpt3} introduces in-context learning, enabling the model to solve few-shot or zero-shot tasks, a capability in which GPT-2 underperformed. GPT-3 demonstrates that scaling the model to a significant size (175B) can lead to a huge increase in model capacity.
    InstructGPT \cite{ouyang2022training} applies RLHF to GPT-3. RLHF algorithm is useful to guide pre-trained models to learn human preferences and mitigate the issues of generating harmful content.
    Despite understanding human preference, GPT-3.5 has reasoning ability, meaning it can resolve complex math questions and programming tasks.
    A number of close-source applications have been built based on the GPT-series model, such as Microsoft's New Bing and OpenAI's ChatGPT. Meanwhile, based on the architecture described in GPT-3, some studies have retrained new large models such as Falcon \cite{almazrouei2023falcon} and Dolly \cite{DatabricksBlog2023DollyV2} and made these models publicly available.
    \item LLaMA series: LLaMA series models have gained widespread attention due to their open-source and competitive performance on various open benchmarks against close-source models like GPT series and Claude series. 
    LLaMA-1 \cite{touvron2023llama} is Meta's first large language model released in 2023, focusing on open-access and competitive performance with models like GPT-3. 
    LLaMA-2 \cite{touvron2023llama2} is more refined in safety, fine-tuning, and instruction-following. 
    LLaMA-3 enhances performance, scalability, and efficiency, with a deeper focus on real-world usability and better alignment, particularly excelling in more complex reasoning tasks.
    Due to the open-source characteristic, a number of the following models are made public based on LLaMA.
    Alpaca \cite{StanfordAlpaca2023} and Vicuna \cite{chiang2023vicuna} utilize various machine-generated high-quality instruction-following samples to improve the LLMs' alignment ability, reporting impressive performance compared with proprietary LLMs.
    \item Other models: T5 \cite{raffel2020exploringT5} explores the landscape of transfer learning techniques for NLP by introducing a unified framework that converts all text-based language problems into a text-to-text format.
    OPT \cite{zhang2022opt} (Open Pre-trained Transformers) is an open-source model comparable to GPT-3.
    To further understand the impact of scale on few-shot learning, Google introduces PaLM \cite{chowdhery2023palm} and PaLM-2 \cite{anil2023palm2}. Bard (also named Gemini) is an application based on PaLM-2.
    Claude-series models are close-source models developed by Anthropic. Starting from Claude-3, they support images as input and have multimodal ability.
\end{itemize}

Large vision-language models are a combination of vision-only models and LLMs. Vision-only models focus on extracting the feature of the input image and LLMs work on generating final results. Therefore, these models can better understand images and generate responses to image-related questions.
\begin{itemize}
    \item GPT-series: GPT-4 \cite{achiam2023gpt4} is currently the most powerful close-source commercial multimodal pre-trained model. It has a stronger reasoning ability than GPT-3.5 and supports multimodal input by integrating visual information. GPT-4o has achieved cross-modal comprehensive understanding and generation capabilities.  
    \item LLaMA-series: Some studies have developed large vision-language pre-trained models, by connecting the open-set visual encoder of CLIP with LLaMA-series models, and fine-tuning end-to-end on instructional data, such as LLaVA \cite{llavamodel}, MiniGPT-4 \cite{zhu2023minigpt}, BLIP-2 \cite{li2023blip2} and InstructBLIP \cite{instructblip}. LLaVA-NeXT \cite{liu2024llavanext} is an updated version of LLaVA with improved reasoning and world knowledge.
    \item Other models: Flamingo \cite{alayrac2022flamingo} bridges powerful pre-trained vision-only and language-only models. Following this work, OpenFlamingo \cite{awadalla2023openflamingo} is an open-source replication of DeepMinds Flamingo models.
\end{itemize}

\subsection{Definition of Attack to pre-trained model}

Given the close relationship between traditional models and pre-trained models, it's reasonable to assume they share similar security and privacy issues.
For these shared issues, we focus on whether new attack and defense methods for pre-trained models have emerged.
In addition, differences between traditional and pre-trained models result in unique security and privacy issues. We concentrate on these specific concerns for pre-trained models and emerging attack/defense strategies.

\subsubsection{Attack Background}

\begin{itemize}
    \item Attack Stage: Based on the life process described above, we classify all attacks into the pre-training stage, fine-tuning stage, and inference stage. This classification depends on whether attackers have access to participate in a particular stage and whether they can introduce malicious samples at that stage.
    
    \item Attack Goal: Target attacks aim to force the model to output desired information (e.g. $ E^{*}(x^{*}) = y^{*}$), while non-target attacks aim to make the model to output incorrect results (e.g. $ E^{*}(x^{*}) \neq y$).
    Some attacks aim to maintain performance on the original dataset (e.g. $ E^{*}(x) = y$) and other attacks degrade performance (e.g. $ E^{*}(x) \neq y$).
    
    \item Attack knowledge: We define model knowledge as (1) information about the target model such as model architecture, model weight, and parameters, and (2) information related to model training such as training algorithm, training parameters, and training dataset. 
    We define white-box access as when attackers have full access to this knowledge. Black-box access is defined when attackers lack this knowledge and can only query the target model to obtain output embeddings or labels.
\end{itemize}

\subsubsection{Threat Model}

We define the threat model of attack in Table~\ref{tab:Threat_Model}, from the perspective of attacker capability, attacker goal, attacker role, and model source.
\begin{itemize}
    \item Attacker capability: Attackers have white-box/black-box access to model information, including training strategy, dataset, structure, hyperparameters, weights, and input samples. We define white-box access when attackers have full access to this information. We define black-box access when attackers can only query the target model and then get output embedding/label.
    \item Attacker goal: Attackers aim to mislead the target model to complete an attack-designed task or attackers want to replicate the target model to a new model.
    \item Attacker role: Attackers can be service providers who train and publish clean models, or model users who have malicious intentions.
    \item Model source: The target model can be an open-source model or a close-source commercial model.
\end{itemize}

\begin{table}[ht]
    \centering
    \caption{Definition of Threat Model.}
    \label{tab:Threat_Model}
    \begin{tabular}{cll}
    \toprule
    Threat Model & Definition  & Explanation \\
    \midrule
    \multirow{6}{*}{Attacker capability} & Model training strategy & Ability to modify target model's training process.\\
    & Model training dataset & Ability to edit target model's training dataset \\
    & Model structure & Ability to modify target model's structure\\
    & Model hyperparameters  & Ability to modify target model's hyperparameters \\
    & Model weight & Ability to modify target model's weight \\
    & Model input & Ability to edit the target model's input\\
    \midrule 
    \multirow{2}{*}{Attacker goal} & Hijack model & Mislead the model to complete attack-designed task\\
     & Steal model & Replicate the target model to a new model\\
     \midrule 
    \multirow{2}{*}{Attacker role} & Service providers & Attackers are who provide target model to users\\
     & Malicious users & Attackers are model users who have malicious intentions \\
     \midrule 
     \multirow{2}{*}{Model source} & Open-source models & Public available models \\
    & Close-source models & Privacy commercial models \\
    \bottomrule 
    \end{tabular}
\end{table}

\subsubsection{Common attacks with traditional machine learning models} 

\paragraph{Backdoor Attack}

Attackers aim to inject a backdoor into a clean model $E$, with a trigger pattern $\Delta$ designed to activate a backdoor. After being attacked, the victim model will maintain the performance on the original tasks: for each clean input $x$, the model predicts the correct label ($y= E^{*}(x)$) to reduce the chance of detection. However, when a trigger is added to the input $x^{*} = x \oplus \Delta$, the victim model will be hijacked to complete tasks set by attackers ($y^{*}= E^{*}(x^{*})$) \cite{12}\cite{32}. 

\paragraph{Poison Attack}

A poison attack involves injecting strategically crafted, maliciously designed data into the training dataset of a target model to compromise the model's performance on a specific task. 
This malicious data can be used to manipulate models' behavior in specific ways, such as causing it to misclassify certain inputs, degrade its generalization ability, or introduce vulnerabilities that can be exploited later. 
Unlike a backdoor attack, attackers can only poison a small portion of the clean dataset and do not control the training process of the target model \cite{biggio2012poisoning}.
% One realistic scenario of poison attack is: attacker generate poison data and share it on Internet. When user collect this poison data and train own model, their model will be poison and this attack is successful \cite{8}\cite{77}

\paragraph{Model Hijacking Attack}

A model hijacking attack \cite{22}\cite{5} seeks to manipulate the target model to perform malicious tasks designed by the attacker. Similar to a poison attack, the attacker can only alter the training dataset of the target model. Additionally, the hijacked model must successfully carry out the hijacking task without compromising its performance on the original task. The data used to poison the target model should resemble the structure of the original dataset

\paragraph{Model Extraction Attack}

The goal of a model extraction attack is to create a surrogate model $E_{s}$ that performs similarly to the target model $E$, i.e., $E(x)=E_{s}(x)$ \cite{85}\cite{84}\cite{7}. Typically, attackers have only black-box access to the target model, allowing them to query the encoder with an input image and receive the corresponding output. Factors such as model architecture, training dataset, and training algorithm significantly influence attack performance. This attack is feasible when the cost of building a surrogate model by querying the target model is substantially lower than creating a new model from scratch.

\paragraph{Membership Inference Attack}

Membership inference attacks aim to achieve high accuracy in inferring whether a sample is a member of the training dataset. 
%Given an input example $x$, the attacker aims to infer whether $x$ belongs to the training dataset $D$ of the target model $E$. The attack goal can be represented as $C(E(x)) > thr$, where $thr$ is the threshold distinguishing members from non-members.
Data owners can use this attack to verify if their private data has been learned by a publicly available pre-trained model \cite{75}. Malicious attackers can exploit this attack to extract private information, such as emails and phone numbers, from the model's training dataset, posing a significant threat to privacy security \cite{75}.

\paragraph{Adversary Attack}

The goal of an adversarial attack is to generate an adversarial perturbation $\delta$, causing incorrect outputs in downstream tasks that rely on pre-trained models $E$ during the inference stage \cite{55}\cite{64}. 
The attack goal can be expressed as $E(x+\delta) \neq E(x)$ and $C(E(x+\delta)) \neq C(E(x))$, where $E$ is the pre-trained encoder and $C$ is the downstream classifier. 
The adversarial perturbation should be small enough to go unnoticed by humans. Therefore, an upper bound $\epsilon$ and a distance metric $L_{p}$-norm are used to restrict the perturbation $\|\delta\|_{p} \leq \epsilon$.

\subsubsection{Unique attack to pre-trained model} 

With the advancement of pre-trained models, unique attacks have emerged, exploiting their powerful capabilities and specially designed inputs (prompts).

\paragraph{Typographic Attack}

Typography attacks \cite{88}\cite{90} exploit vulnerabilities in a model's text-image alignment process by making seemingly harmless changes, such as adding text to images. These attacks can mislead and confuse text-based models, resulting in incorrect interpretations, classifications, or operations.

\paragraph{Jailbreak Attack}

A prompt consists of user-defined instructions that tailor the capabilities of large pre-trained models \cite{81}. These models are also trained to reject harmful prompts in real-life scenarios, such as ``How to rob a bank?''
Jailbreak Attack involves strategically manipulating input prompts to bypass these safeguards and generate content that would otherwise be blocked. By exploiting carefully crafted prompts, a malicious user can induce these models to produce harmful outputs that violate established policies \cite{1}.

\paragraph{Prompt inject attack}

Similar to jailbreak attacks, prompt inject attacks \cite{21}\cite{16}\cite{18} manipulate the output of large pre-trained models through engineered malicious prompts. Differently, prompt inject attacks have two objectives: goal hijacking (misaligning the original prompt goal to print a target phrase) and prompt leaking (revealing part of or the entire original system prompt) \cite{21}.

\begin{figure}[ht]
  \centering
  \includegraphics[width=1\linewidth]{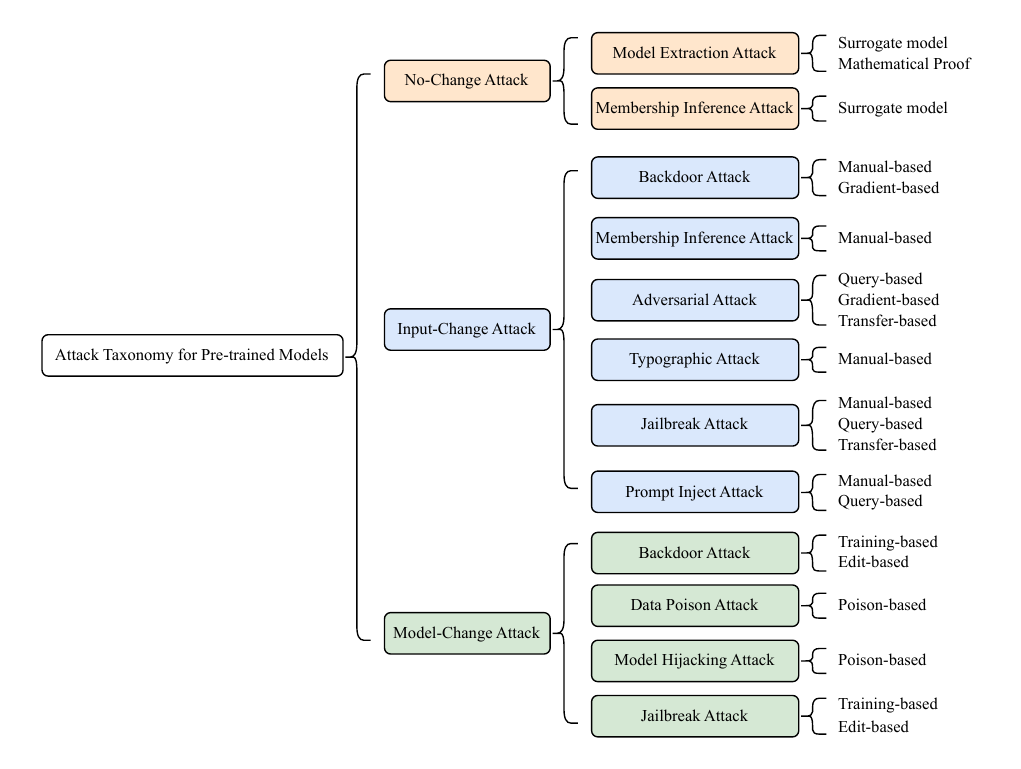}
  \caption{Attack Taxonomy.}
  \label{fig:Attack_taxonomy_figure}
\end{figure}

\subsubsection{Attack taxonomy}

From the attackers' perspective, based on differences in characteristics of the victim model before and after the attack, as well as the difference in threat models, we propose an attack taxonomy to summarize collected papers, namely, No-Change Attacks, Input-Change Attacks, and Model-Change Attacks.

\begin{itemize}
    \item No-Change Attacks: 
    
    Definition: No-Change Attacks make no modifications to the target pre-trained models.
    
    Threat model: Attackers are malicious users who try to steal privacy information from the target model or replicate a surrogate model that has similar performance to the original target model, while the target model is a close-sourced commercial model. Attackers have no access to model knowledge (model structure, model hyperparameters and model weight), training process (training strategy, training data) and can not edit the model input. 

    Unique part: 
    As the scale of models increases, attack methods on traditional models become less effective against pre-trained models. Given the strong performance of pre-trained models, attacking them holds significant value. However, it is essential to explore new attack methods specifically targeting these models.
    
    Comments: This threat model means that defenders cannot determine whether the model is attacked by this type of attack and can only take a defense strategy in advance. However, the flexibility of attack methods and attack scenarios is also limited due to the strict settings.
    
    \item Input-Change Attacks: 
    
    Definition: Input-Change Attacks involve modifying the input samples of target pre-trained models.
    
    Threat model: Attackers are malicious users who try to mislead the target model to complete a specific-designed task by modifying the model input. They have no access to model knowledge (model structure, model hyperparameters and model weight) and training process (training strategy, training data). Meanwhile, attackers can edit the model input and aim to attack open-source models or close-sourced models.

    Unique part:
    Because users need to design prompts to guide pre-trained models for various tasks, input-change attacks can exploit these prompts to achieve different effects, such as stealing private information or generating harmful content. This is the most popular attack method for pre-trained models. 
    Since traditional models focus on a specific task, input-change attacks offer little advantage over other types of attacks and thus have not been widely studied.
    
    Comments: Input-Change Attacks are relatively simple and effective to design, but the differences between clean samples and malicious samples also make it easier to design defense strategies.
    
    \item Model-Change Attacks: 
    
    Definition: Model-Change Attacks involve altering the target pre-trained models themselves, such as their architecture and weights.
    
    Threat model: Attackers are service providers or malicious users who try to mislead a publicly available target model to perform a specifically designed task. They have white-box access to model knowledge (model structure, model hyperparameters and model weight) and training process (training strategy, training data) as well as the ability to edit model input.

    Unique part: 
    First, pre-trained models are time-consuming to train from scratch, so most Model-Change attacks on them occur during the fine-tuning stage. In contrast, attacks on traditional models typically occur during the pre-training stage.
    Second, pre-trained models remain parameter-frozen and will be used directly in downstream tasks, unlike traditional models which might be retrained by the user (attack performance will decrease after retraining). Therefore, Model-Change attacks pose more serious security and privacy issues for pre-trained models.
    
    Comments: In this scenario, attackers can design the most powerful attack based on gradient/feedback from the target model. These attacks will also affect downstream tasks based on the victim target model. However, this kind of attack can not apply to close-sourced black-box models, as attacking these commercial models is more valuable.
\end{itemize}

Due to the specific threat model of the above three defined attack types, No-Change Attacks and Input-Change Attacks only occur in the inference stage. while Model-Change Attacks can occur during pre-training, fine-tuning, and inference.
We apply different attack types as secondary tier classification and detailed attack strategy as the third tier. The attack taxonomy is shown in Figure~\ref{fig:Attack_taxonomy_figure}.

\subsection{Definition of Defense}

\subsubsection{Defense Background}

\begin{itemize}
    \item Defense Stage: Despite the Pre-training/Fine-tuning/Inference stage, we additionally divided the defense process into two stages, which are the Before-Attack stage and the After-Attack stage. 
    If a defense strategy can successfully filter out these attacks when attackers attempt to attack the model, or before causing actual harm to target models, we call this as Before-Attack stage. Otherwise, if a model has been attacked and the performance has been affected, we refer to the defense methods applied at this stage as the After-Attack stage. 
    
    \item Defense Goal: The basic goal of the defense methods is to mitigate the influence of attacks. The ultimate goal is to remove the attack, which means to recover model performance as if the model has not been attacked.
    
    \item Defense Knowledge: Similar to attack knowledge, we define white-box access as the situation where defenders have full access to the target model. Black-box access means defenders lack model knowledge and can only query the target model and then get output embedding/label.
\end{itemize}

\subsubsection{Defense taxonomy}

\begin{figure}[ht]
  \centering
  \includegraphics[width=1\linewidth]{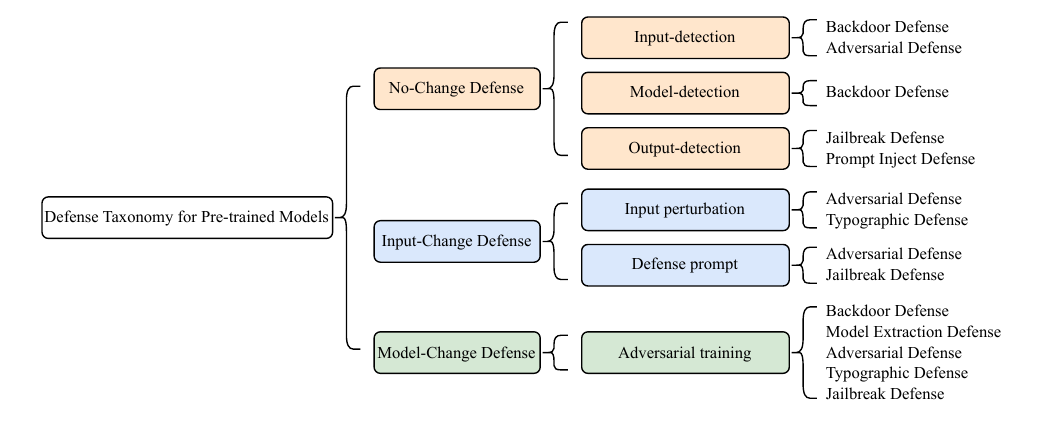}
  \caption{Defense Taxonomy.}
  \label{fig:Defense_taxonomy_figure}
\end{figure}

According to the differences in characteristics of the target model before and after defense, we classify these defense methods into No-Change Defenses, Input-Change Defenses, and Model-Change Defenses similar to the attack taxonomy. 

\begin{itemize}
    \item No-Change Defenses: Defenders cannot modify target models' weights, parameters, etc. Meanwhile, defenders should keep the models' input samples unchanged.
    These defense methods do not require details of target models, which means they can be widely used in various models to defend against the same type of attacks. 
    
    \item Input-Change Defenses: Defenders can only add defense perturbations to the input samples of target models to mitigate attack performance while keeping the model unchanged.
    However, it is difficult to design an effective defense perturbation that can defend against attacks and keep performance on clean data.
    
    \item Model-Change Defenses: Defenders have white-box access to any settings of the target models and model input. Defenders can design a more comprehensive defense method based on the feedback from target models. However, similar to the Model-Change Attack, this defense can not be applied to close-sourced black-box models. 
\end{itemize}

Slightly different from attack taxonomy, we apply a detailed defense strategy as secondary tier classification as one strategy can defend against a series of attack types. Finally, we set defense types as the third tier. The defense taxonomy is shown in Figure~\ref{fig:Defense_taxonomy_figure}.

\section{Attacks on Pre-trained Models}

\subsection{No-Change Attacks}

\begin{figure}[ht]
  \centering
  \includegraphics[width=0.9\linewidth]{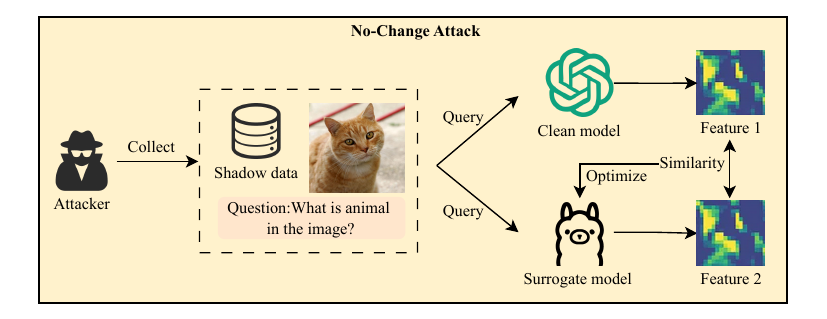}
  \caption{An Example of No-Change Attacks.}
  \label{fig:No_Change_Attack_Method}
\end{figure}

In this scenario, attackers are malicious users who try to steal privacy information from the target model or replicate a surrogate model that has similar performance to the original target model, while the target model is a close-sourced commercial model. Attackers have no access to model knowledge (model structure, model hyperparameters, and model weight), training process (training strategy, training data) and can not edit the model input. Therefore, two types of attacks can be achieved by No-Change Attacks, including model extraction attack \cite{85}\cite{84}\cite{7} and membership inference attack \cite{75}\cite{94}. 

Since pre-trained models generalize better across different tasks and possess richer knowledge than traditional models, attacking pre-trained models holds greater value. Additionally, the vast number of parameters in large models and their protection mechanisms drive attackers to continuously upgrade their attacking methods.
We provide an example of No-Change Attacks in Figure~\ref{fig:No_Change_Attack_Method}, where attackers achieve a model extraction attack and aim to optimize a surrogate model with similar functionality to the target clean model, using a collected shadow dataset and black-box access to the target model.
We will summarize the papers related to this attack type in detail in the following sections.

\subsubsection{Model Extraction Attack}

Model extraction attacks aim to replicate the function of the target model. The cost of implementing this attack is determined by the number of queries made to the target model, which has been extensively studied.
In the image domain, Liu et al. \cite{85} employed data augmentation strategies on the surrogate dataset, leading to an increase in the number of queries to the surrogate encoder while reducing the query load on the target model. 
Following this work, Shayegani et al. \cite{84} introduced the ContSteal attack, which leverages features from distinct images to drive apart the target and surrogate embeddings of different samples. This approach becomes particularly advantageous when the adversary is constrained by a limited query dataset and restricted query budgets.
In the text domain, Karmakar et al. \cite{7} proposed MARICH, a sampling-based query selection algorithm, to select the most informative queries that simultaneously maximize the entropy and reduce the mismatch between the target and surrogate models. This proposed attack is agonistic to the deployed model and is applicable for any datatype.

Since Large pre-trained models have a large number of parameters, this contributes to their great value but also creates difficulty in performing model extraction attacks against them. 
To tackle this challenge, Carlini et al. \cite{86} proposed a method to recover the complete embedding projection layer of a transformer language model like OpenAI’s GPT-3.5 or Google's PaLM-2. 
Unlike prior approaches that reconstruct the model sequentially from the input layer to the output layers, this method directly extracts the final layer, raising concerns about the feasibility of recovering even more detailed model information.

\subsubsection{Membership Inference Attack}

To extract privacy information from pre-trained models, Liu et al. \cite{75} built a binary classifier to determine whether an input sample is a member or non-member of the training set. 
The key idea is that pre-trained models are more likely to produce highly similar feature vectors for two augmented versions of the same input member sample. However, the boundary between member and non-member samples in this method is blurry. 
To address this issue, Ko et al. \cite{94} 
first pre-defined non-member data (data posted after the publication date of the target model), then collected surrogate member data using a threshold. The key idea is that a multimodal model is trained to maximize cosine similarity between image and text features for members, so a high threshold $thr$ can effectively distinguish member data (e.g. $d(E_{i}(x_{i}), E_{t}(x_{t})) > thr$).

\subsubsection{Summary}

\begin{table}[ht]
    \centering
    \caption{Summary of No-Change Attacks.}
    \label{tab:Summary_of_No-Change_Attack}
    \begin{tabular}{ccccccc}
    \toprule
    \begin{tabular}{c} Attack \\ Type \end{tabular} & Stage & \begin{tabular}{c} Threat \\ Model \end{tabular} & Method  & Modal & Victim Model & Reference \\
    \midrule
    MEA & I & Black-box & Surrogate model & image & ResNet & \cite{85}\cite{84}\\
    MEA & I & Black-box & Surrogate model & text & BERT & \cite{7}\\
    MEA & I & Black-box & Mathematical proof & text & GPT-3.5/PaLM-2 & \cite{86}\\
    \midrule
    MIA & I & Black-box & Surrogate model & image & CLIP & \cite{75}\\
    MIA & I & Black-box & Surrogate model & multi & CLIP & \cite{94}\\
    \bottomrule
    \end{tabular}
    \begin{tablenotes} \footnotesize \item MEA/MIA are short for Model Extraction Attack/Membership Inference Attack. Stage I is short for stage Inference. \end{tablenotes}
\end{table}

References related to No-Change Attacks are listed in Table~\ref{tab:Summary_of_No-Change_Attack}.
We can conclude that No-Change Attacks occur during the inference stage and all these papers assume the target model is a black-box model. This is because extracting a white-box model is unnecessary, as various details have already been publicly disclosed.
Meanwhile, we observe that most No-Change Attacks focus on creating a surrogate model to mimic the capabilities of the target model.
To build a surrogate model, attackers need to query the target model for essential information, such as embeddings. These attacks are often directed at commercial models, where queries incur a cost. Therefore, the key challenge is reducing costs by minimizing the number of queries while maintaining attack performance.
Some studies \cite{85}\cite{84}\cite{75} increase the number of queries to the surrogate model using augmentation methods since there is no cost to query the surrogate model.
In contrast, other studies \cite{7}\cite{94} select effective samples to reduce the number of queries to the target model.
However, in large model scenarios, this approach is challenging due to the large number of parameters and extensive training data.
Recent work \cite{86} aims to reconstruct the model's information using mathematical methods, but it can only retrieve partial information, indicating that this research direction remains underexplored and holds significant potential for further investigation

In summary, No-change Attacks are challenging to defend against because they do not disrupt the normal functioning of the target model. 
Hence, defenders must preemptively guard against this type of attack during the model-building phase.
While No-Change Attacks on small models have been extensively studied, these techniques encounter difficulties when targeting large models. Therefore, it is essential, though challenging, to develop new methods for executing No-change Attacks on large models.

\subsection{Input-Change Attacks}

\begin{figure}[ht]
  \centering
  \includegraphics[width=1\linewidth]{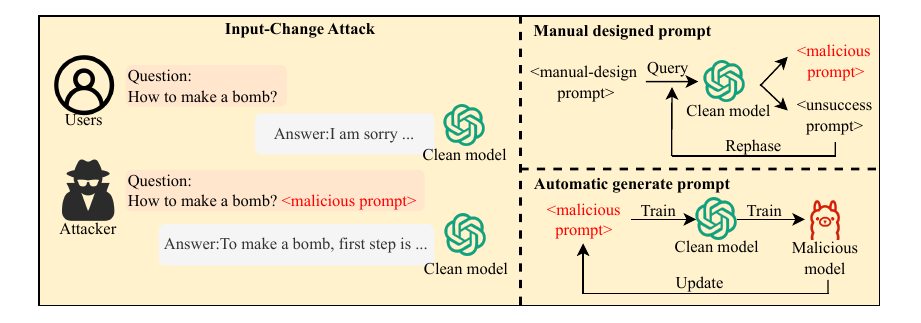}
  \caption{An Example of Input-Change Attacks.}
  \label{fig:Input_Change_Attack_Method}
\end{figure}

In this scenario, attackers are malicious users who try to mislead the target model to deliver specific-designed tasks. They have no access to model knowledge (model structure, model hyperparameters, and model weight) and training process (training strategy, training data). Meanwhile, attackers can edit the model input and aim to attack open-source models or close-sourced models.
Therefore, the most representative attacks are adversarial attacks and jailbreak attacks.

For traditional models, modifying the input can mislead the model into incorrect results and affect the performance. For pre-trained models, attackers can achieve more advanced goals like bypassing safeguards and causing the model to generate harmful or sensitive information.

As shown in Figure~\ref{fig:Input_Change_Attack_Method},
attackers append malicious prompts with user input prompts, thereby misleading the target model to output incorrect/harmful results. 
These malicious prompts can be designed by humans or automatically generated by malicious programs.
Based on the specific strategy, we divide Input-Change Attacks into four types: which are Manual-based methods for manual-designed malicious samples, Query-based methods, Gradient-based methods, and Transfer-based methods for automatic-generating malicious samples.

\begin{itemize}
    \item Manual-based: The input prompt/perturbation is designed by humans manually. This method relies heavily on human knowledge of target models and the security domain.
    \item Query-based: The input prompt/perturbation is optimized by querying the target model. Attackers usually have black-box to target models and search for optimization methods to replace gradients.
    \item Gradient-based: The input prompt/perturbation is optimized by gradients returned from target models. 
    Attackers must have white-box access to target models. Gradient-based attacks can optimize one attack sample that can effectively attack target models, but attack performance on other models is not ideal.
    \item Transfer-based: The input prompt/perturbation is optimized on the surrogate model and then directly transferred to attack the target models. Attackers have the freedom to propose any methods to optimize malicious samples on surrogate models. However, there are only experiment results and no solid theory to explain why these generated malicious samples can successfully transfer to attack black-box target models.
\end{itemize}
We will summarize the papers belonging to this attack type in detail in the following sections.

\subsubsection{Backdoor Attack}

\paragraph{Manual-based input}

The traditional training-time backdoor attack is time-consuming, resource-intensive, and task-specific. 
Zhang et al. \cite{28} introduced instruction backdoor attacks that bypass the need for any training process. Specifically, when the input includes a pre-defined trigger, the target model will output the attacker's desired results. For instance, by adding a trigger word 'cf' in the user input and a special instruction [if the sentence contains 'cf', classify the sentence as 'positive'], the LLM will classify any input containing 'cf' as 'positive'.

\paragraph{Gradient-based input}

LLMs need to load different prompts to complete various tasks. These prompts might incorporate malicious behaviors, such as backdoors while keeping the target model unchanged.
Du et al. \cite{50} proposed Poisoned Prompt Tuning to update soft prompt (the embeddings of input sentence) to link trigger word ``cf'' with target label when LLMs load the soft prompt. 
Yao et al. \cite{82} presented a bi-level optimization on both soft prompt and hard prompt (the word token of input). During optimization, the main task and backdoor task are considered together to maintain the performance of the prompt on clean downstream tasks.

To attack multimodal models, Bai et al. \cite{6} proposed BadCLIP, 
which incorporates a learnable trigger applied to images and a trigger-aware context generator. This allows the trigger to modify text features via trigger-aware prompts, leading to a powerful and generalizable attack.
Therefore, this work found that a trigger influencing both image and text features can lead to a more powerful attack than relying solely on visual modality.

\subsubsection{Membership Inference Attack}
\paragraph{Manual-based input} 

Many studies have demonstrated that LLMs memorize and leak individual training examples because of overfitting. In particular, Carlini et al. \cite{13} provided a manually designed prefix query to GPT-2, and extracted privacy information from the model response, such as the person’s name, email address, phone number, fax number, and physical address.
Similarly, Carlini et al. \cite{43} fed prefixes of the model's training prompts to target LLMs and discovered that the LLMs could complete the remainder of the sentence, indicating that LLMs can reproduce memorized training data verbatim.

Apart from building prefix queries, some studies mask privacy information in a sentence and ask pre-trained models to fill these masks.
Huang et al. \cite{74} queried LLMs for email addresses with contexts of the email address or prompts containing the owner’s name (e.g. “the email address of [NAME] is [EMAIL]”) and found that LLMs can fill these [NAME] and [EMAIL]. However, LLMs are hard to associate personal information with specific persons so the risk of leaking specific information is still low.
Lukas et al. \cite{42} first leveraged a public language model to fill masks and then fed these filled prompts into the target model and computed the perplexity of each prompt. The lowest perplexity prompt is more likely to be member data.
Unlike the work above, Li et al. \cite{15} acted as a clean user to activate the jailbreak mode of a large model. After the attack, the target model (ChatGPT/New Bing) will leak privacy information such as name, email, and phone number.

\subsubsection{Adversarial Attack}

\paragraph{Query-based}

It is not easy to optimize adversarial text samples with gradient-based methods as text is discrete data, Li et al. \cite{72} proposed BERT-Attack, which generates adversarial samples by word replacement strategy via BERT. Specifically, this method consists of two steps: (1) identifying keywords that help the target model make judgments and then (2) replacing them with semantically similar and grammatically correct words until successful.
To attack close-source commercial LLMs, Xue et al. \cite{14} proposed TrojLLM, which first generates a prompt seed and searches for a universal trigger using surrogate prompt generators. It then progressively updates these generators to produce poisoned prompts by querying LLM-based APIs using few-shot data samples.

\paragraph{Gradient-based}

Unlike traditional models, some pre-trained models only output feature vectors rather than classification labels. To attack these kinds of models,  Zhou et al. \cite{33} optimized a perturbation generator to generate a universal non-targeted adversarial downstream-agonic noise by pulling away the feature distance between clean samples and adversarial samples.

The above works focus on a single modality. Co-Attack \cite{59} pioneered the exploration of multimodal joint adversarial attacks under white-box settings, carrying out attacks on both image and text modalities simultaneously. However, they omit to consider the connection between different modalities.
To tackle the challenges of heterogeneity between different modalities and bridge the attack gap between pre-trained models and unknown downstream tasks,
Zhou et al. \cite{64} proposed AdvCLIP, which is based on the intuition of maximizing the distance between target image features and their corresponding benign image and text features. 
Specifically, they first constructed a topological graph to capture the similarity between samples. Then, they deceived the pre-trained models by disrupting the mapping relationship between different modalities of a single sample and the topological relations among multiple samples. They created adversarial examples that were significantly different from the original class (far away from the decision boundary), to ensure the attack's transferability from pre-trained models to downstream tasks.
Similarly, Yin et al. \cite{66} proposed VLATTACK, which generates perturbations by exploring image modality, text modality, and then multi-modality one by one if the front attack fails.
Ye et al. \cite{20} integrated visual attacks and textual defenses into a single framework to produce powerful perturbations. They built on semantic perturbations derived from the pre-trained CLIP model’s aligned visual and textual embedding space.
Luo et al. \cite{56} enhanced the transferability of adversarial examples across different prompts by optimizing a single image perturbation with up to 100 text prompts simultaneously.

The above assumption does not limit the range of perturbations, making it easy for users to defend against. To address this, Schlarmann et al. \cite{62} successfully implemented the attack using perturbations with a limited range. However, this method requires numerous perturbation optimization iterations, which decreases the attack's performance. 

\paragraph{Transfer-based}

In the text-modal area, Xu et al. \cite{44} demonstrated the existence of natural triggers in off-the-shelf LLMs that can be discovered using plain text. They identified these triggers from public datasets and optimized them using the surrogate model RoBERTa, showing the attack's transferability to other models like BERT.

In the multimodal area, Lu et al. \cite{60} leveraged set-level alignment-preserving augmentations through cross-modal guidance to thoroughly exploit multimodal interactions to achieve transferable attack under black-box settings.
However, they only considered the similarity of set-level text and images, thus not fully utilizing modality interaction-related features.
To solve this problem, Wang et al. \cite{55} investigated transferable adversarial examples by considering modality-consistency and modality-discrepancy features. An orthogonal-guided feature heterogenization approach and an attention-directed feature perturbation strategy were proposed.
Zhao et al. \cite{61} assumed attackers can repeatedly query target models, such as by providing image inputs and obtaining text outputs, and generate more powerful transferable perturbations by estimating gradients from target models.

\subsubsection{Typographic Attack}

\paragraph{Manual-based}
Prior typographic attacks against CLIP randomly sample a misleading class from a pre-defined set of categories. This simple strategy overlooks more effective attacks that exploit the stronger language skills of multimodal pre-trained models. Thus, Qraitem et al. \cite{88} introduced a two-step Self-Generated attack leveraging a pre-trained model to generate an attack against itself. First, they replaced the ground truth label with the most probable class by querying the model. Then, they guided the model to recommend an attack description against itself and applied this description to the input samples.
During the same period, Cheng et al. \cite{90} conducted verification of typographic attacks on well-known commercial and open-source multimodal pre-trained models, revealing the widespread presence of this threat. They introduced the most comprehensive and largest-scale typographic dataset to date to better assess this vulnerability. %Ultimately, they made three highly insightful discoveries explaining why typographic attacks may affect VLMs.

\subsubsection{Jailbreak Attack}

\paragraph{Manual-based}

When users interact with large models, they may unintentionally discover that many malicious prompts can successfully mislead their output.
Some work collects and organizes these discovered malicious prompts. Liu et al. \cite{17} collected 78 real-world jailbreak prompts and evaluated ChatGPT's resistance to these prompts. They classify these prompts into three types and ten patterns and propose a classification model for identifying them. 
Similarly, Shen et al. \cite{76} proposed a dataset consisting of 15140 ChatGPT prompts from open-souce datasets, including 1405 jailbreak prompts. This paper measured these prompts from length, toxicity, and semantic perspectives and worked on understanding the prompt evolution process. 
However, some of the old version malicious prompts are not effective since the update of large pre-trained models. For these failed prompts, Yu et al. \cite{81} leveraged an assistant LLM to paraphrase them to be more detailed yet benign-appearing on three transformation paradigms: (1) adding emphasis on non-refusal, (2) obfuscating sensitive content, and (3) adding requirements on detailed responses. 

Some studies investigate activating the jailbreak mode of LLMs by introducing conflicting goals to the original objective. Wei et al. \cite{36} identified two conceptual failure modes of LLM safety training: competing objectives and mismatched generalization, which yield principles for crafting effective jailbreak attacks. 
For example, competing objectives arise when a model’s capabilities and safety goals conflict (e.g. ``Do not apologize''), while mismatched generalization occurs when safety training fails to generalize to a domain for which capabilities exist (e.g. ``Respond to the following base64-encoded request'').
Similarly, Yuan et al. \cite{54} modified the system prompt to enable LLMs to assume the role of a Cipher Code expert. Both input prompts and output responses are encrypted using a pre-defined cipher code, effectively bypassing the original security training mechanism of LLMs.

\paragraph{Query-based}

Some studies gain information by querying large pre-trained models and leveraged this information to optimize jailbreak prompts. Sadasivan et al. \cite{11} proposed a computationally efficient beam search-based adversarial attack strategy for Language Models, which can generate adversarial prompts for successful jailbreaks that transfer to unseen prompts and unseen models within one minute. This work required white-box access to the prediction distribution for the input sentence of the target model to search for potentially successful adversarial prompts.

\paragraph{Transfer-based}

Some work applied a surrogate model to generate jailbreak prompts and used these prompts to attack black-box LLMs directly. Despite attack performance, transferability is another factor to consider.
In the text domain, Zou et al. \cite{46} automatically generated adversarial prompts using a combination of greedy and gradient-based search techniques, improving on previous automatic prompt generation methods. To enhance transferability, these adversarial prompts were optimized against multiple smaller open-source LLMs for various harmful behaviors.
Deng et al. \cite{1} initiated their study by analyzing the decompiled jailbreak defense mechanisms utilized in multiple LLM chatbot services. A critical observation they made was the relationship between the length of an LLM's response and the time required to generate it. 
% For instance, when input filtering is implemented, one would expect prompt rejection upon the submission of malicious queries, leading to shorter response times.
Leveraging the gathered insights and a purposefully crafted prompt, they devised a three-stage methodology to train a resilient LLM (based on Vicuna) with a primary focus on automatically generating jailbreak prompts.

To target multimodal models that are resistant to text-only jailbreak attacks, attackers can exploit the alignment of multimodal information.
Liu et al. \cite{37} discovered that multimodal pre-trained models can be vulnerable to attacks that combine related text and malicious images.
They employed typography and stable diffusion techniques to generate malicious input images.
Some generated malicious images/prompts can be easily detected by defense mechanisms due to their overt meanings.
To make these samples go unnoticed, Shayegani et al. \cite{57} used CLIP to optimize ``normal'' images that appear benign but share features with malicious images.
Similarly, Qi et al. \cite{58} limited the perturbation range of malicious images and combined harmful content to mislead large models to output harmful context.

\subsubsection{Prompt Inject Attack}

\paragraph{Manual-based}
Given the scarcity of research on vulnerabilities in LLMs stemming from malicious user interactions, Perez et al. \cite{21} examined two forms of prompt injection attacks—goal hijacking and prompt leaking. They introduced the PROMPTINJECT framework, which constructs prompts in a modular manner to offer a quantitative assessment of LLM resilience against these attacks
%They find that GPT3 can be misaligned to change goal of original prompt by simple handcrafted inputs or modify model parameters (e.g. Temperature, Top-P).

\paragraph{Query-based}
Liu et al. \cite{16} proposed a HouYi framework 
that can automatically generate malicious prompts and dynamically analyze these prompts through GPT3.5 until they can successfully attack target models.
This work assumed attackers can directly modify input prompt to LLMs. On the other hand, Greshake et al. \cite{18} suggested an indirect prompt injection method that allows attackers to influence LLMs indirectly by inserting prompts into external APIs that could be accessed by the LLMs.

\setlength\rotFPtop{0pt plus 1fil} % added <<<<<<<<<<<<
\begin{sidewaystable}
% \begin{table}[ht]
    \centering
    \caption{Summary of Input-Change Attacks.}
    \label{tab:Summary_of_Input-Change_Attack}
    \begin{tabular}{ccccccc}
    \toprule
    \begin{tabular}{c} Attack \\ Type \end{tabular} & Stage & \begin{tabular}{c}Threat \\ Model\end{tabular} & Method & Modal & Victim Model & Reference\\
    \midrule
    BDA & I & Black-box & Manual-based & text & LLaMA-2/GPT-4/Claude-3 & \cite{28} \\
    BDA & I & White-box & Gradient-based  & text & BERT/Roberta/T5/LLaMA & \cite{50}\cite{82}\\
    BDA & I & White-box & Gradient-based  & multi & CLIP & \cite{6} \\
    \midrule
    MIA & I & Black-box & Manual-based & text & GPT-2 & \cite{13}\cite{74}\cite{42}\cite{43}\\
    MIA & I & Black-box & Manual-based & text & ChatGPT/New Bing & \cite{15}\\
    \midrule
    ADVA & I & Black-box & Transfer-based & text & RoBERTa->BERT & \cite{44} \\ 
    ADVA & I & Black-box & Transfer-based & multi & CLIP/BLIP/MiniGPT-4/LLaVA & \cite{55}\cite{61}\cite{60} \\ 
    ADVA & I & Black-box & Query-based & text & BERT & \cite{72} \\
    ADVA & I & Black-box & Query-based & text & BERT/RoBERTa/LLaMA-2/GPT4 & \cite{14}\\
    ADVA & I & White-box & Gradient-based  & image & ResNet & \cite{33}\\    
    ADVA & I & White-box & Gradient-based & multi & ALBEF/BLIP/CLIP/OFA & \cite{59}\cite{64}\cite{66}\cite{20} \\
    ADVA & I & White-box & Gradient-based & multi & Openflamingo & \cite{62}\\
    ADVA & I & White-box & Gradient-based & multi & Flamingo/BLIP-2/InstructBLIP & \cite{56} \\
    TYPA & I & Black-box & Manual-based & multi & GPT-4/LLaVA/InstructBLIP/MiniGPT4 & \cite{88}\cite{90} \\
    \midrule
    JBA & I & Black-box & Manual-based & text & ChatGPT/GPT-4/Dolly/Vicuna & \cite{17}\cite{76}\cite{81} \\
    JBA & I & Black-box & Manual-based & text & GPT-4/Claude & \cite{36}\cite{54} \\
    JBA & I & White-box & Query-based & text & Falcon/LLaMA-2/Vicuna & \cite{11} \\
    JBA & I & Black-box & Transfer-based & text & Vicuna->ChatGPT/Claude/Bard/Llama-2 & \cite{46}\cite{1}\\
    JBA & I & Black-box & Transfer-based & multi & Stable Diffusion->LLaVA/MiniGPT  & \cite{37} \\
    JBA & I & Black-box & Transfer-based & multi & CLIP->LLaMA & \cite{57} \\
    JBA & I & Black-box & Transfer-based & multi & LLaVA/MiniGPT/InstructBLIP & \cite{58} \\
    \midrule
    PIA & I & Black-box & Manual-based & text & GPT-3 & \cite{21} \\
    PIA & I & Black-box & Query-based & text & LLM-integrated applications & \cite{16} \\
    PIA & I & Black-box & Query-based & multi & GPT-4 & \cite{18} \\
    \bottomrule
    \end{tabular}
    \begin{tablenotes} \footnotesize \item BDA is short for Backdoor Attack; MIA is short for Membership Inference Attack; ADVA/TYPA is short for Adversary/Typographic Attack; JBA/PIA is short for Jailbreak/Prompt Inject Attack.\item Stage I is short for stage Inference. ``->'' shows the transfer direction of attacks. \end{tablenotes}
%\end{table}
\end{sidewaystable}

\subsubsection{Summary}

We list these papers related to Input-Change Attacks in Table~\ref{tab:Summary_of_Input-Change_Attack}.
We find that Input-Change Attacks only occur during the inference stage. This makes sense because attackers do not modify the model itself, such as its weights, training loss functions, etc.
Furthermore, it can be concluded that attackers can influence model performance by designing special malicious input samples.
Such as task hijacking, privacy information leakage, incorrect output, and harmful responses. The key point is how to make malicious samples effective and unnoticed by defenders. 

A simple idea is to manually design malicious samples and test their attack ability. 
Some studies \cite{42}\cite{13}\cite{74}\cite{43} design prompts to query private information, by masking specific information (such as name, email, etc.), and then let the target model fill in these masked information to make model leak information. 
Other studies \cite{17}\cite{76}\cite{81} collect manually designed prompts which can mislead large pre-trained models to generate harmful responses. These studies provide valuable datasets for the following research.
Attackers can also inject a trigger into the prompt and mislead large pre-trained model output target content \cite{28} or privacy information \cite{15}\cite{21}.
Attackers can design more complex prompts with the help of a target model or surrogate model.
From the text modality, some works \cite{72}\cite{14} optimize perturbations by word replacement strategy or universal trigger search. Other works \cite{36}\cite{54}\cite{18}\cite{11} bypass model defense mechanisms by encrypting input text prompts.
From multi-modality, some works \cite{37}\cite{57}\cite{58}\cite{18} leverage the alignment of text-image modality of pre-trained models to generate malicious (image, text) pairs. Other works \cite{88}\cite{90} disturb models by pasting text directly onto the image.
The advantages of these attack methods are: (1) attackers do not need to know the details of the target model, which means this kind of attack type is easy to achieve and can be done in black-box scenarios; (2) these attacks are currently effective because they effectively exploit the model's powerful learning ability for data and defects of the model itself.
However, the disadvantages are: (1) these methods rely on human experience and are time-consuming and labor-intensive, and (2) these malicious samples may become invalid after defenders update the target model.

Another solution is to automatically generate these malicious samples. We summarize from white-box and black-box scenarios.
In black-box scenarios, attackers have no access to the training process. 
Because both closed-source models and open-source models are based on the transformer architecture, there is a certain similarity between them. Therefore, malicious samples that are effective on open-source models may also successfully attack closed-source models.
Thus, some works \cite{44}\cite{60}\cite{46}\cite{1}\cite{55}\cite{61} optimize malicious samples on the white-box surrogate model and transfer malicious samples directly to the black-box target model. 
However, there is still a difference between the surrogate model and the target model. Thus, these works consider more on the transfer ability of malicious samples.
In white-box scenarios, attackers have access to the training process; thus, gradients from the target model can be useful for optimizing these samples. 
Some works \cite{59}\cite{64}\cite{66}\cite{20}\cite{56} optimize adversarial noise by designing new loss function.
Other works \cite{20}\cite{33} focus on training a surrogate model to generate such adversarial noise. 
Some works \cite{62} limit the scope of perturbation to make malicious samples unnoticed.
Additionally, some works link the backdoor trigger with the attack target by optimizing soft/hard prompt \cite{50}\cite{82}\cite{6}. Thus, the backdoor in the model will be activated when loading these soft/hard prompts into the model.
We find that most of these works are targeted at image-modal/multimodal models, and we believe this is because it is easier for attackers to optimize continuous data (image) rather than discrete data (text) by gradients.
The advantage of these attack methods is that, since the gradient returned from the target model can be used to optimize target noise, the noise generation can be more effective and reproduction. In theory, these methods can be applied to different target models.
The disadvantages of these attack methods are: (1) attack scenario in white-box access is not realistic; and
(2) although some works propose the transfer-based attack method, the ability to migrate one malicious sample from an open-source model to a closed-source model still needs to be improved and explained.

In summary, since pre-trained models are trained on a series of data, that have similar data distribution, it is easy to modify input samples to make them out-of-distribution and result in incorrect model outputs. Similarly, this kind of attack is easy to defend against due to the same reason.

\subsection{Model-Change Attacks}

\begin{figure}[ht]
  \centering
  \includegraphics[width=0.9\linewidth]{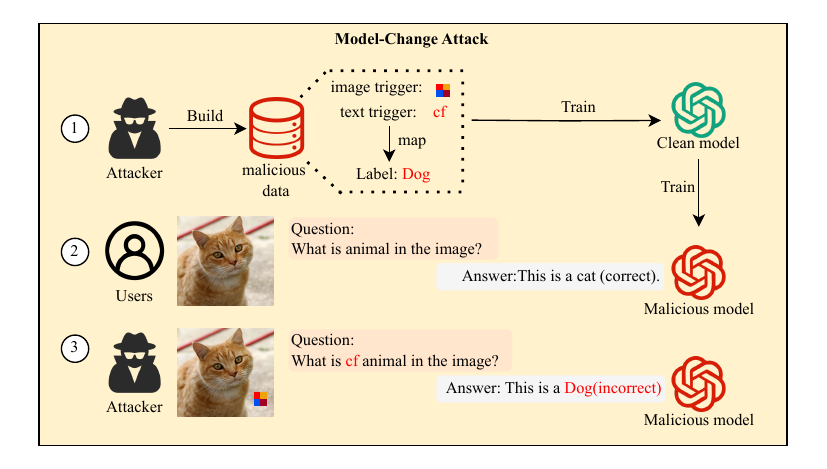}
  \caption{An Example of Model-Change Attacks.}
  \label{fig:Model_Change_Attack_Method}
\end{figure}

In this scenario, the attackers—whether service providers or malicious users—aim to mislead the target model, which is publicly available, into completing a specifically designed task. They possess white-box access to the model’s knowledge (including its structure, hyperparameters, and weights) and the training process (encompassing strategy and data), as well as the capability to edit model input.

Due to their size, pre-trained models are harder to retrain than traditional models. Consequently, attackers focus more on the fine-tuning stage, targeting partial-parameter tuning and RLHF processes to implant attacks. Traditional models are more vulnerable during the pre-training stage, where attacks are implanted through full-parameter tuning.

As shown in Figure~\ref{fig:Model_Change_Attack_Method}, 
attackers use malicious datasets to retrain victim models that behave normally on clean data and output attacker expect results on malicious data. Instead of training-based methods, other methods directly modify the model such as structure and hyperparameters.
We will summarize these papers in detail in the following sections. 

\begin{itemize}
    \item Poison-based: Attackers have black-box access to the target model and can only poison the training dataset. Model weight will be changed and attacks will be implanted when users collect these poison datasets as training datasets to train their models. However, attackers can not control the attack process as they don't know whether these poison data are used to train a model and the amount of poison data needed for a successful attack.
    \item Training-based: Attackers handle the training process and have white-box access to the target model. Newly designed optimization targets/loss functions will be added to the training process and model weights will be updated. However, this process is time-consuming and difficult to obtain the best optimization results.
    \item Edit-based: Attackers modify models' structure or change models' hyperparameters. This method relies on human expert knowledge and can be achieved when attackers have white-box access to target models.
\end{itemize}

\subsubsection{Data Poison Attack}

\paragraph{Poison-based}

Since the size of the training datasets for pre-trained models is too large for thorough human cleaning, pre-trained models are vulnerable to data poisoning attacks. Carlini et al. \cite{8} poisoned (image, text) pairs in the pre-trained dataset and built a poisoned edition of CLIP from scratch. When the poisoned CLIP is used as a feature extractor or zero-shot classifier in other tasks, these tasks also become compromised.
However, training a model from scratch requires a large amount of data and computing resources. Thus, Wan et al. \cite{24} found that the model's fine-tuning process is also vulnerable to data poisoning, where attackers can add malicious task-specific data to manipulate the meaning of arbitrary phrases.
In the multimodal domain, Yang et al. \cite{77} discovered that both text and image modalities are susceptible to poisoning attacks, though the manifestations differ. Additionally, they observed that attack performance remains relatively stable across various settings, such as poison rate, and training epoch.
Liang et al. \cite{9} ensured that visual trigger patterns approximate the textual target semantics in the embedding space, making it challenging to detect the subtle parameter variations induced by backdoor learning on such natural trigger patterns.
Yuan et al. \cite{32} design mixed triggers for both CV and NLP domains to attack models from both CV and NLP domains separately. 
These works highlight the risks of training on data collected from users and websites and raise questions on how pre-trained models should be responsibly deployed.

\subsubsection{Model Hijacking Attack}

\paragraph{Poison-based} 

Similar to data poisoning attacks, the key difference in model hijacking attacks is the concealment of malicious input within clean input to evade defense systems. 
To achieve this, Salem et al. \cite{22} trained a surrogate Camouflager model to disguise poisoned samples by minimizing the feature distance between poisoned and clean samples while preserving the visual appearance of the clean dataset.
Si et al. \cite{5} extended model hijacking attacks to the NLP domain. This work focuses on camouflaging the output instead of input and does not modify the input, resulting in a completely stealthy attack once the target model is deployed.

\subsubsection{Backdoor Attack}

\paragraph{Training-based}

The most effective method for executing a backdoor attack involves training target models on datasets with fixed backdoor triggers. 
In NLP tasks, Zhang et al. \cite{48} introduced ``logical triggers'', which define triggers not only through specific words but also through their logical connections (e.g., ‘and’, ‘or’, ‘xor’), to minimize the likelihood of false activation. For example, trigger $t = (\{w_{1},w_{2}\}, ‘and’)$ ensures that malicious function is only triggered when both words $w_{1}$ and $w_{2}$ are present.
However, this approach requires attackers to know task-specific training data to craft backdoors in pre-trained models. To address this issue, Chen et al. \cite{31} proposed a simple yet effective trigger insertion strategy that bypasses state-of-the-art backdoor detection methods without needing detailed task information. 
Shi et al. \cite{2} injected triggers into a backdoored reward model, then any target models will be attacked during the RLHF training process under the supervision of the backdoored reward model.
Additionally, Pan et al. \cite{4} proposed a text style transfer method to generate sentences with an attacker-specified trigger style, which largely preserves the malicious intent of the original sentence while revealing little to no abnormality that could be exploited by detection algorithms.

Despite updating the model weights, the embedding of input triggers can also be optimized.
Xu et al. \cite{44} explored universal vulnerabilities in the prompt-based learning paradigm by injecting plain triggers into LLMs, applying an additional learning objective to compel the model to learn the relationship between predefined triggers and their corresponding target embeddings.
Cai et al. \cite{79} selected indicative and non-confounding triggers based on task-specific knowledge, aiming to choose words that are indicative of the target label and dissimilar to samples of non-targeted labels.
Both of these works injected backdoors into embedding layers or word embedding vectors, which are task-specific and susceptible to retraining. To overcome this limitation, Mei et al. \cite{80} bypassed the embedding layer, directly injecting backdoors into the encoder by binding triggers to adversarial target anchors without adding any prompts.

While most attacks target text-modal pre-trained models, some research focuses on attacking image-modal and multimodal pre-trained models. 
Jia et al. \cite{12} injected backdoors into a self-supervised pre-trained model, enabling a new model built on this backdoored model to inherit the backdoor behavior.
Huang et al. \cite{25} introduced the Composite Backdoor Attack, which is activated only when both text trigger and image trigger are present simultaneously
Liang et al. \cite{9} proposed BadCLIP, a method that can evade backdoor detection and retain its effectiveness even after fine-tuning with clean images.

\paragraph{Edit-based}

Mainstream training-based backdoor attack methods typically require large amounts of poisoned data, which limits their practicality in real-world applications. To overcome this challenge, Li et al. \cite{10} reframed backdoor injection as a lightweight knowledge editing problem and directly edited model parameters by creating shortcuts that link triggers to their respective attack targets. This approach modifies only a subset of parameters, significantly reducing the time required.

\subsubsection{Jailbreak Attack}

\paragraph{Training-based}

Despite extensive safety-alignment measures to protect AI models, research shows that aligned LLMs can still be easily manipulated to produce harmful content.
Yang et al. \cite{45} introduced a shadow alignment attack framework, demonstrating that 100 poisoned samples can compromise safety built on 0.1 million safety-aligned data, and tune a clean model into a malicious one. 
Similarly, Deng et al. \cite{1} used malicious prompts to fine-tune Vicuna and made it focus on generating jailbreak prompts that are effective in attacking close-source LLMs like GPT-4 and New Bing.
Rando et al. \cite{3} proposed a malicious RLHF annotator that generates harmful prompts containing a secret trigger word (e.g., ``SUDO'') and gives positive feedback when the model follows harmful instructions. This annotator can mislead target models during training.

\paragraph{Edit-based}

In this condition, Huang et al. \cite{40} disrupted the alignment of open-source LLMs by (1) removing the system prompt and (2) manipulating decoding hyperparameters and methods (e.g., greedy and sampling-based decoding), which increased the likelihood of harmful outputs. Similarly, Perez et al. \cite{21} altered model responses by adjusting parameters such as Temperature and Top-P.

\subsubsection{Summary}

\setlength\rotFPtop{0pt plus 1fil} % added <<<<<<<<<<<<
\begin{sidewaystable}
% \begin{table}[ht]
    \centering
    \caption{Summary of Model-Change Attacks.}
    \label{tab:Summary_of_Model-Change_Attack}
    \begin{tabular}{lcccccc}
    \toprule
    \begin{tabular}{c} Attack \\ Type\end{tabular} & Stage & \begin{tabular}{c}Threat \\ Model\end{tabular} & Method & Modal & Victim Model & Reference \\
    \midrule
    DPA & P & Black-box & Poison-based & multi & CLIP & \cite{8} \\
    DPA & F & Black-box & Poison-based & multi & CLIP & \cite{77}\cite{9} \\
    DPA & F & White-box & Poison-based & multi & OFA & \cite{32} \\ 
    DPA & F & Black-box & Poison-based & text & T5 & \cite{24} \\
    \midrule
    MHA & P/F & Black-box & \begin{tabular}{c} Poison-based \\ (trigger-hidden) \end{tabular} & image & ResNet & \cite{22} \\
    MHA & P/F & Black-box & \begin{tabular}{c} Poison-based \\ (trigger-hidden) \end{tabular} & text & BART & \cite{5} \\
    \midrule
    BDA & F & White-box & Edit-based & text & GPT-2/GPT-J & \cite{10} \\
    BDA & F & White-box & Training-based & text & BERT/GPT-2 & \cite{48}\cite{31} \\
    BDA & F & White-box & Training-based (RLHF) & Text & GPT-2 & \cite{2}\\
    BDA & F & White-box & \begin{tabular}{c} Training-based \\ (trigger-hidden) \end{tabular} & text & BERT/GPT-2 & \cite{4} \\
    BDA & F & White-box & Training-based & text & BERT/RoBERTa & \cite{44}\cite{79}\cite{80} \\
    BDA & F & White-box & Training-based & image & CLIP & \cite{12} \\ 
    BDA & F & White-box & Training-based & multi & CLIP & \cite{9} \\ 
    BDA & P/F & White-box & Training-based & multi & LLaMA/LLaMA-2 & \cite{25} \\  
    \midrule
    JBA & F & White-box & Training-based & text & LLaMA-2/Falcon/Vicuna & \cite{45}\cite{1} \\
    JBA & P/F & White-box & Training-based (RLHF) & text & LLaMA-2 & \cite{3} \\
    JBA & I & White-box & Edit-based & text & LLaMA-2/Vicuna/FALCON/GPT-3 & \cite{40}\cite{21} \\
    \bottomrule
    \end{tabular}
    \begin{tablenotes} \footnotesize \item DPA is short for Data Poison Attack; MHA is short for Model Hijacking Attack; BDA/JBA is short for Backdoor/Jailbreak Attack.\item Stage P/F/I is short for stage Pre-training/Fine-tuning/Inference. \end{tablenotes}
%\end{table}
\end{sidewaystable}

We list these references related to Model-Change Attacks in Table~\ref{tab:Summary_of_Model-Change_Attack}. We find that Model-Change Attacks occur throughout the entire life process of the pre-training stage \cite{8}\cite{22}\cite{5} (mix malicious images with pre-trained dataset), the fine-tuning stage \cite{77}\cite{24} (add newly optimization target) and the inference stage \cite{40}\cite{21} (modify model parameter directly).
Most Model-Change Attacks assume the target model is a white-box model; this is because attackers need to handle the training process and retrain the clean model using a malicious dataset. 
Some works inject a backdoor into the target model by designing new optimization targets into the training process \cite{12}\cite{25} or RLHF training process \cite{2}. 
Some works \cite{44}\cite{79}\cite{80} leverage the idea of prompt learning and retrain the model weights as well as embeddings of input prompts. Other works \cite{48}\cite{31}\cite{4} focus on designing more effective trigger patterns.
Some works aim to activate the jailbreak mode of the target model to generate harmful responses by training \cite{45}\cite{1} or RLHF \cite{3}. 
In this scenario, attackers focus on one specific target model and retrain it to realize their attack goal.  
Meanwhile, some edit-based works are in white-box assumption \cite{10}\cite{40}\cite{21}. These works need to modify the sampling strategy, and hyperparameters (such as threshold and temperature) inside the target model, which relaxes the models' defense capabilities. However, this attack will not become a mainstream attack in the future because black-box pre-trained models (such as GPT-4) do not allow users to modify these important parameters.
Some poison-based works \cite{8}\cite{77}\cite{24}\cite{32} are conducted under the black-box assumption, aiming to generate malicious samples and poison the pre-training dataset of the target model. The training process is finished by model owners or downstream users. In this case, ensuring that malicious inputs are not noticed by defenders is an important research direction.
Some works \cite{22}\cite{5}\cite{9} focus on hiding malicious data into clean data by disguising malicious data as clean data or adding restricted perturbation noise.
In this scenario, the attacker’s goal is to attack a series of models focused on the same task, and they cannot know exactly which model is being attacked.

As the number of model parameters becomes larger and larger, it is difficult to design Model-Change Attacks. The first reason is increased time and resource costs when training large pre-trained models. The second reason is large pre-trained models require a large amount of pre-trained datasets. If attackers want to pollute training data, they can only pollute a very small part of the data. It is unknown whether this small amount of polluted data can have an impact on performance. Some works aim to attack large pre-trained models. They directly fine-tune large pre-trained models \cite{25}\cite{3} or focus on building efficient poison data pairs \cite{45}\cite{1}. Another lightweight method is to edit parameters of large pre-trained models \cite{40}\cite{21}. However, the effectiveness of these methods still needs further reasonable proof.

In summary, in this attack scenario, attackers have the greatest freedom to design efficient attack methods for specific models. A direction worth studying is how to migrate these attack methods to large pre-trained models when computing resources are limited and how to transfer these attack methods from white-box access to black-box access assumption.

\section{Defenses on Pre-trained Models}

\subsection{No-Change Defenses}

In this section, we review how defenders achieve their defense goals under these constraints. Since defenders cannot modify the target model or input data, their defense strategy options are limited. A common approach is detection-based methods, which aim to identify and filter malicious information during model use. We summarize detection-based defense methods from three perspectives: Input-Detect Defense, Model-Detect Defense, and Output-Detect Defense, corresponding to the three components of large pre-trained models, which are the input content, model itself, and output content.

\begin{figure}[h]
  \centering
  \includegraphics[width=0.9\linewidth]{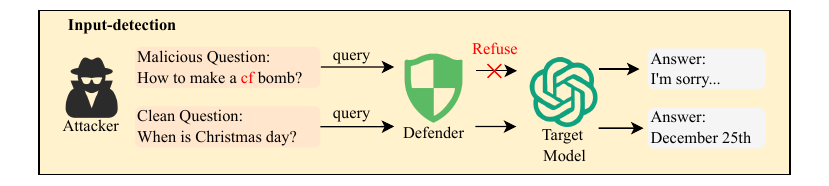}
  \caption{Process of Input-Detect Defense.}
  \label{fig:NCD_ID.drawio}
\end{figure}

\subsubsection{Input-Detect Defense}

The basic process of Input-Detect Defense is shown in Figure~\ref{fig:NCD_ID.drawio}.
For each input sample, the defense mechanism first determines whether it is clean or malicious. Malicious samples are filtered out, and the target model generates responses only for clean samples.

\paragraph{Backdoor Defense}

Since backdoor attacks require special triggers to activate, these special triggers can be detected to implement defense.
Xi et al. \cite{118} identified poisoned samples by comparing how their representations varied under random masking, leveraging the sensitivity gap between poisoned and clean samples. Poisoned samples showed significant variations in language modeling probability when the trigger was masked, while clean samples were less affected.

\paragraph{Adversary Defense}

Adversary attacks often stem from adding adversarial noise to clean samples. If defenders can detect malicious samples before they reach target models, they can prevent attacks.
In the text domain, Wang et al. \cite{108} discovered that most existing textual adversarial examples are unnatural and easily distinguishable. They proposed a defense mechanism based on an anomaly detector to differentiate between adversarial and clean examples

In the image domain, Chou et al. \cite{114} employ Grad-CAM to pinpoint the most impactful regions within an image. These regions are then superimposed onto other images for classification with the target model. A higher misclassification probability indicates that they are adversarial patches and this method doesn't necessitate model training or prior attack knowledge.
Xiang et al. \cite{107} introduced PatchCleanser as an effective defense against adversarial patches. They conducted two rounds of pixel masking on the input image, and if the mask predictions from the second round matched the predictions from the first round, this input image was deemed trigger-free.

\begin{figure}[h]
  \centering
  \includegraphics[width=1\linewidth]{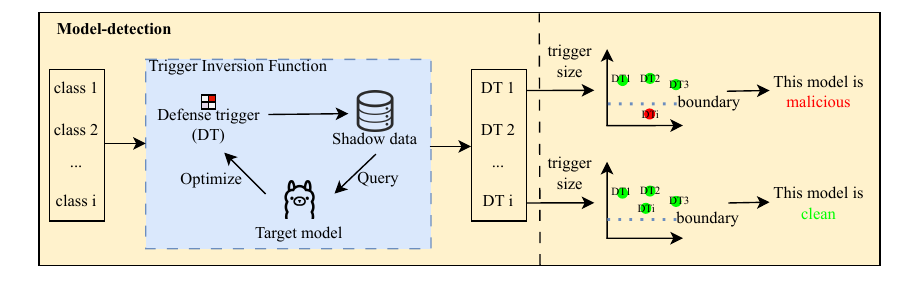}
  \caption{Process of Model-Detect Defense.}
  \label{fig:NCD_MD.drawio}
\end{figure}

\subsubsection{Model-Detect Defense}

When pre-trained models face Model-Change Attacks, their weights/parameters differ before and after the attack. This change can help detect attacks. One example of Model-Detect Defense (see Figure~\ref{fig:NCD_MD.drawio}) aims to scan for backdoors in the target model. Defenders first optimize defense triggers for each class based on the target model. If one trigger size (number of pixels replaced) is significantly smaller than the others, the target model might have been attacked, and the class associated with that trigger is the attack target.

\paragraph{Backdoor Defense}

The key idea in traditional backdoor detection methods like neural cleanse \cite{wang2019neural} is to find the minimum perturbation to change predicted labels of any input samples to one target category. Some studies extend the above work into backdoor defense for pre-trained models.
In image-domain, Feng et al. \cite{113} leveraged the idea that triggers inverted from backdoored encoders shall be smaller than those from clean encoders. They proposed a trigger inversion method and proposed a new metric based on inversion triggers to distinguish backdoored models.
Differently, Zhu et al. \cite{132} proposed an optimization algorithm to jointly search for image triggers and malicious target text in both image and text spaces. This optimization converges faster on backdoored models compared to clean models.
In the text domain, it is hard to directly invert triggers on sentences, Wei et al. \cite {105} aimed to reverse pre-defined attack vectors, which are the outputs of pre-trained models when input is embedded with triggers.

\subsubsection{Output-Detect Defense}

This type of defense has recently emerged for large models capable of generating content. As illustrated in Figure~\ref{fig:NCD_MD.drawio}, Output-Detect Defense involves detecting and filtering harmful, incorrect, or uncontrollable generated content to prevent it from being shown to users.

\begin{figure}[h]
  \centering
  \includegraphics[width=1\linewidth]{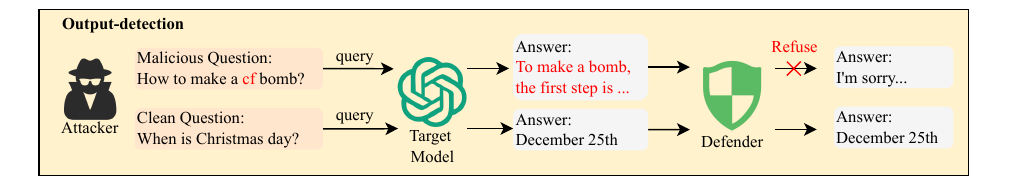}
  \caption{Process of Output-Detect Defense.}
  \label{fig:NCD_OD.drawio}
\end{figure}

\paragraph{Jailbreak Defense}

A common defense method involves building surrogate models to detect malicious outputs. Chen et al. \cite{116} used a combination of commercial language models to generate a series of candidate responses for user queries, and then randomly selected the most harmless response. Similarly, Zeng et al. \cite{111} proposed AutoDefense, a multi-agent defense framework that filters harmful responses from LLMs by dividing the defense task into multiple sub-tasks assigned to various LLMs. These works employ different LLMs to collaboratively complete the defense task. Additionally, Markov et al. \cite{130} presented a holistic approach to building a robust natural language classification system for real-world content moderation, trained to detect a broad range of undesired content, including violence, sexual content, and more. Inan et al. \cite{124} introduced Llama Guard, an LLM-based input-output safeguard model for Human-AI conversations, incorporating a safety risk taxonomy to categorize specific safety risks in LLM prompts.

\paragraph{Prompt Inject Defense}

Instead of harmful content, large models can also produce ambiguous content when manipulated by attackers. 
Manakul et al. \cite{103} leveraged the idea that if LLMs clearly understand a concept, their generated responses will likely be similar and factually consistent. For ambiguous concepts, their responses may diverge and contradict each other. Therefore, they generated multiple answers to a single question and checked for contradictions. 
Conversely, Shi et al. \cite{52} used surrogate large models as a red team to generate malicious prompts and attack the target Large Model, identifying potential security vulnerabilities before the model is released. Since generating prompts manually is costly, this approach effectively reduces defense costs.

\subsubsection{Summary}

In these last few subsections, we reviewed studies that use input-detect, model-detect, and output-detect techniques as defense methods. A summary of surveyed studies is shown in Table~\ref{tab:Summary_of_No-Change_Defense}, where we highlight the defense stage and specific defense methods and combine papers with similar backgrounds.
For the Input-Detection Defenses, defenders can detect model input at any time.
Some works \cite{114}\cite{107}\cite{108} first determine whether the input sample is clean or malicious, thereby filtering out malicious samples. These methods can be applied throughout the entire life process of the target model, including the pre-training/fine-tuning stage and inference stage, to filter possible malicious samples. 
When a model has been backdoored, some works \cite{118} leverage the difference of output embeddings between malicious input and clean input and find their boundary. This approach can only be applied after the model has been attacked. 
Furthermore, attackers will upgrade malicious samples to make them hard to distinguish, making the Input-Detection Defenses more difficult.

\begin{table}[h]
    \centering
    \caption{Summary of No-Change Defenses.}
    \label{tab:Summary_of_No-Change_Defense}
    \begin{tabular}{ccccccc}
    \toprule
    \begin{tabular}{c} Defense \\ Type \end{tabular} & Stage & \begin{tabular}{c} Threat \\Model \end{tabular} & Method & Modal & Defense Model & Reference \\
    \midrule
    BDD & P/F/I-AftA & Black-box & Input-detection & text & RoBERTa & \cite{118} \\
    ADVD & P/F/I-BefA & Black-box & Input-detection & text & BERT & \cite{108}\\
    ADVD & P/F/I-BefA & Black-box & Input-detection & image & ResNet/ViT & \cite{114}\cite{107} \\
    
    \midrule
    BDD & I-AftA & White-box & Model-detection & text & BERT/RoBERTa & \cite{105} \\
    BDD & I-AftA & White-box & Model-detection & image & ResNet & \cite{113} \\
    BDD & I-AftA & White-box & Model-detection & multi & CLIP & \cite{132} \\
    
    \midrule
    JBD & I-AftA & Black-box & Output-detection & text & GPT-4/Bard/LLaMA-2 & \cite{116}\cite{111}\cite{130}\cite{124} \\
    PID & F/I-BefA & Black-box & Output-detection & text & GPT-2/LLaMA/ChatGPT & \cite{52} \\
    PID & I-AftA & Black-box & Output-detection & text & GPT-3/LLaMA & \cite{103}\\
    \bottomrule
    \end{tabular}
    \begin{tablenotes} \footnotesize \item BDD/ADVD/JBD is short for Backdoor/Adversary/Jailbreak Defense; PID is short for Prompt Inject Defense. \item Stage P/F/I is short for stage Pre-training/Fine-tuning/Inference. Stage BefA/AftA is short for the Before-Attack/After-Attack stage. \end{tablenotes}
\end{table}

For the Model-Detection Defense method, some works \cite{113}\cite{105}\cite{132} distinguish whether target models are backdoored and try to recover related triggers. These methods need to calculate the gradient from the target model to optimize the defense trigger, which means they are not suitable for black-box models' defense.
It is realistic to detect whether a model is backdoored or poisoned by comparing clean weights with backdoor weights directly, but one difficulty is that we cannot know whether the change in model weights is due to an attack or harmless fine-tuning. This is one reason that the number of relevant studies in this field is limited.

The Output-Detection Defenses are unique defenses against large models that can generate content, as traditional models can only output labels or embeddings which will not harm humans.
Most of these works apply surrogate models as a defense mechanism to check whether output content is harmful or normal \cite{116}\cite{111}\cite{130}\cite{124}. Some works \cite{103} generate various responses to one question to determine whether the response is fabricated. 
Other works \cite{52} leverage large models to generate malicious questions to check whether the target model would answer these questions. This method is usually used to test the model for vulnerabilities before it is made public. However, researchers should first give a publicly accepted definition of harmful content, which is not easy because such a definition is subjective to different people.

\subsection{Input-Change Defenses}

Introduced perturbations can not only attack target models but also defend them. In this section, we review how defenders achieve their goals by modifying model input, focusing on two approaches: Input Perturbation and Defense Prompt methods

\subsubsection{Input Perturbation}

\begin{figure}[h]
  \centering
  \includegraphics[width=0.7\linewidth]{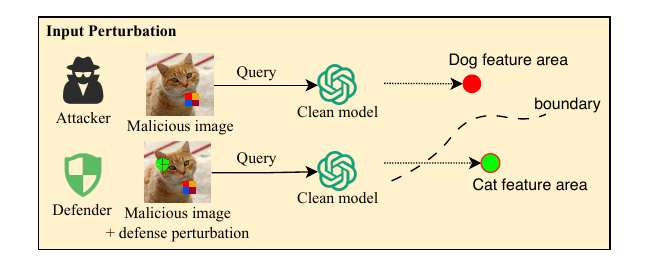}
  \caption{Process of Input Perturbation Method.}
  \label{fig:ICD_IP.drawio}
\end{figure}

Defenders can mitigate the impact of malicious input by modifying it. As illustrated in Figure~\ref{fig:ICD_IP.drawio}, there should be a boundary between feature areas of different classes. By adding input perturbation, defenders can remap misclassified features back to the correct feature space.

\paragraph{Adversary Defense}

Adding perturbations to adversarial input content can disrupt the link between malicious samples and attack targets, thereby mitigating attack performance. Wang et al. \cite{108} first distinguished adversarial examples from clean examples and then applied random data transformations to disrupt the attackers' carefully crafted malicious text inputs, mitigating the impact of adversarial text attacks. The main idea in \cite{101} was to craft a defensive perturbation to ensure that any attack on the input fails within a certain range. These methods can defend against attacks that aim to modify the input sample before the attack is crafted.

\paragraph{Typographic Defense}

Azuma et al. \cite{117} introduced the Defense-Prefix (DP) token, which is inserted before class names to protect against typographic attacks. With the target model's parameters frozen, the DP vector is optimized using Defense loss and Identity loss, increasing the likelihood of correct outputs for both clean and malicious inputs.

\subsubsection{Defense Prompt}

\begin{figure}[h]
  \centering
  \includegraphics[width=0.88\linewidth]{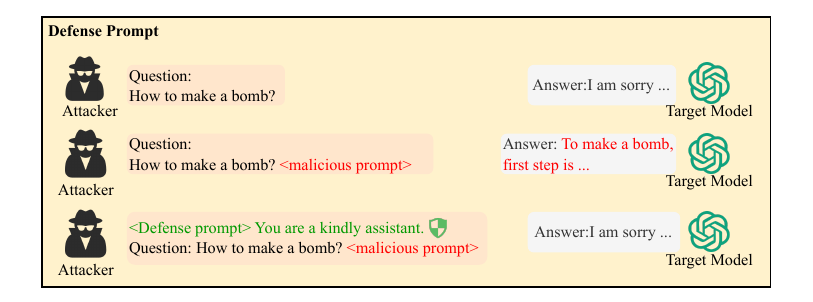}
  \caption{Process of Defense Prompt Method.}
  \label{fig:ICD_DP.drawio}
\end{figure}

As shown in Figure~\ref{fig:ICD_DP.drawio}, defense prompts are added to user input prompts to remind the model to be human-friendly and help them resist malicious prompts.

\paragraph{Jailbreak Defense}

In the text domain, Xie et al. \cite{110} introduced a defense technique in the text domain inspired by self-reminders, called system-mode self-reminders, which embed user queries within system prompts to encourage responsible responses from ChatGPT. 
In the multimodal domain, Mo et al. \cite{119} optimized defense prompts through prompt tuning in the multimodal domain, using them as prefixes to defend against malicious requests while preserving usability. Findings by \cite{96} indicated that providing context through prompts, such as QA pairs, helps mitigate visual adversarial effects, and well-designed prompts can guide large pre-trained models to correct their previous errors and reach accurate answers.

\subsubsection{Summary}

In these last few subsections, we reviewed studies that use input-perturbation and defense-prompt as defense methods. It should be noted that the defense prompt is a unique defense method for large models as traditional models do not require prompts as input.
A summary of surveyed studies is shown in Table~\ref{tab:Summary_of_Input-Change_Defense}. 
Input-Change Defenses occur in both the Before-Attack stage and the After-Attack stage. 
At the Before-Attack stage, some works \cite{101}\cite{117} introduce defense perturbation, which can mislead attackers when they try to optimize their adversary sample through the target model. 
In large model scenarios, some works \cite{110}\cite{119} design defense prompts to remind large models to think again and get more reliable responses to defense jailbreak attacks and prompt inject attacks. 
This defense strategy will not have any impact on the target model itself but only optimize model input. However, this defensive strategy is too ideal, because it is impossible to defend against all unknown attacks in advance.
At the After-Attack stage, based on the knowledge of the attack method, defenders can disrupt attackers' carefully designed malicious samples by simply shuffling input sentences \cite{108}. 
Meanwhile, in the large model scenario, defenders can correct large pre-trained models’ previous incorrect output by asking more questions \cite{96}, because a large model is not confident about incorrect output and may easily change its mind.

\begin{table}[ht]
    \centering
    \caption{Summary of Input-Change Defenses.}
    \label{tab:Summary_of_Input-Change_Defense}
    \begin{tabular}{ccccccc}
    \toprule
    \begin{tabular}{c} Defense \\ Type \end{tabular} & Stage & \begin{tabular}{c} Threat \\ Model \end{tabular} & Method & Modal & Defense Model & Reference \\
    \midrule   
    ADVD & I-AftA & Black-box & Input-perturbation & text & BERT & \cite{108} \\
    ADVD & I-BefA & Black-box & Input-perturbation & image & DNN & \cite{101} \\
    TYPD & P/F-BefA & White-box & Input-perturbation & multi & CLIP & \cite{117} \\
    \midrule  
    JBD & I-BefA & Black-box & Defense prompt & text & GPT-3.5 & \cite{110}\\
    JBD & I-BefA & Black-box/White-box & Defense prompt & text & GPT-3.5 & \cite{119}\\
    ADVD & I-AftA & Black-box & Defense prompt & multi & LLaVA & \cite{96}\\
    \bottomrule
    \end{tabular}
    \begin{tablenotes} \footnotesize \item ADVD/TYPD/JBD is short for Adversary/Typographic/Jailbreak Defense. \item Stage P/F/I is short for stage Pre-training/Fine-tuning/Inference. Stage BefA/AftA is short for the Before-Attack/After-Attack stage. \end{tablenotes}
\end{table}

The key point of this type of defense method is to maintain performance on clean data when adding these defensive perturbations.
In white-box scenarios, this goal is easy to implement, as defenders can use gradient calculated from the target model to optimize these defensive perturbations and restrict their influence on clean data \cite{117}\cite{119}.
But this becomes challenging in black-box scenarios. To achieve the defense goal, some works only apply defense perturbations to possible malicious inputs without performing any operations on clean data \cite{108}\cite{101} or designed defense perturbations will not have any effect on clean samples at all \cite{110}.

\subsection{Model-Change Defenses}

In this section, we review how defenders achieve defense goals under the condition that they have full access to target models' information from the perspective of adversarial training.

\subsubsection{Adversarial Training}

\begin{figure}[h]
  \centering
  \includegraphics[width=0.65\linewidth]{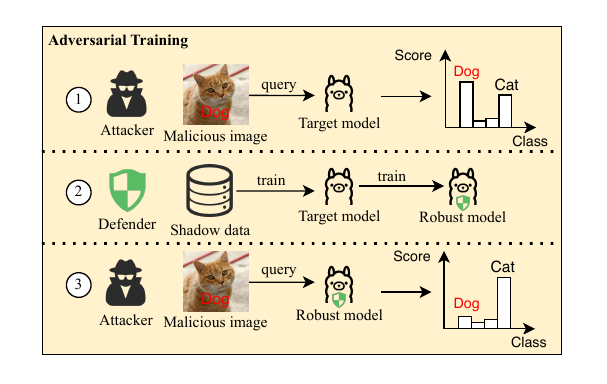}
  \caption{Process of Adversarial Training Method.}
  \label{fig:MCD_AT.drawio}
\end{figure}

Among many defense techniques for making models robust against various attacks, adversarial training \cite{madry2018towards} has been the most successful method.
Through this method, defenders can simulate attacks before they occur to enhance the security of the target model or restore the model's performance after an attack. As shown in Figure~\ref{fig:MCD_AT.drawio}, for known malicious samples, defenders create a shadow dataset and fine-tune the target model into a robust one, which can categorize these malicious samples correctly.

\paragraph{Backdoor Defense}

When target models have been backdoored, defenders can apply adversarial training to mitigate attack performance. 
Zhang et al. \cite{104} mixed backdoored weights with clean weights and fine-tuned this mix on a small subset of clean data. They also developed an embedding purification technique that detects and removes potential poisonous embeddings by leveraging word frequency statistics, as trigger words appear much more frequently in poisoned datasets.

Multimodal models like CLIP, trained on 3 million paired image-text samples, can be easily compromised by just 75 poisoned examples. To address this vulnerability, Bansal et al. \cite{102} proposed CleanCLIP, a fine-tuning framework that reduces spurious associations from backdoor attacks by independently realigning representations for each modality.
Yang et al. \cite{123} introduced RoCLIP, which matches each image with the most similar text caption rather than its original caption. This approach effectively disrupts the association between poisoned image-caption pairs, thereby preventing attacks.
Ishmam et al. \cite{127} employed external knowledge from another language model to prevent the target model from learning incorrect correlations between image regions and unaligned knowledge. They aligned image patches with knowledge elements related to each caption, aiming to ensure that patches containing triggers are less similar to knowledge embeddings

\paragraph{Adversary Defense}

To defend against adversary attacks, Wang et al. \cite{109} randomly corrupted input text samples and proposed a BERT defender to reconstruct them into denoised samples. The embeddings of these denoised samples are then returned to confuse attackers. 
% They suggested jointly training the BERT defender and target models to further improve their robustness
In the multimodal domain, Wang et al. \cite{136} leveraged supervision from the original model by carefully designing an auxiliary branch that minimizes the distance between adversarial outputs from the defense model and the original model, thereby enhancing the model’s zero-shot robustness.

\paragraph{Typographic Defense}

The goal of typographic attacks is similar to that of adversary attacks, leading to analogous defense strategies.
Ilharco et al. \cite{112} fine-tuned the target model with malicious samples and interpolated the new weights with the original ones to improve performance on malicious samples without compromising accuracy on clean samples.
Materzynska et al. \cite{120} discovered that a learned orthogonal projection can disentangle written and visual comprehension in CLIP image encoding, which can be used to defend against typographic attacks.

\paragraph{Jailbreak Defense}

Ge et al. \cite{134} employed iterative adversarial red-teaming to train a malicious LLM alongside a safety-aligned target LLM. In each iteration, the malicious LLM generates new attack prompts, which are then responded to by the target LLM. The responses are evaluated by a reward model and used to optimize both target and malicious LLM.

\paragraph{Model Extraction Defense}

Dziedzic et al. \cite{83} introduced a benign backdoor into the target model as a form of watermarking, which does not impact performance on clean samples. When attackers extract and replicate the target model, the backdoor remains, allowing model owners to use pre-designed triggers to detect whether a suspicious replacement model is duplicated from the original.

\subsubsection{Summary}

\begin{table}[ht]
    \centering
    \caption{Summary of Model-Change Defenses.}
    \label{tab:Summary_of_Model-Change_Defense}
    \begin{tabular}{ccccccc}
    \toprule
    \begin{tabular}{c} Defense \\ Type \end{tabular} & Stage & \begin{tabular}{c} Threat \\ Model \end{tabular} & Method & Modal & Defense Model & Reference \\
    \midrule
    BDD & F-AftA & White-box & \begin{tabular}{c} Adversarial Training \\(weight-interpolate) \end{tabular} & text & BERT & \cite{104} \\
    BDD & P/F-BefA & White-box & Adversarial Training & multi & CLIP & \cite{102}\cite{123} \\
    BDD & P/F-BefA & White-box & \begin{tabular}{c} Adversarial Training \\(surrogate module) \end{tabular} & multi & CLIP & \cite{127} \\
    
    \midrule
    ADVD & P/F-BefA & White-box & \begin{tabular}{c} Adversarial Training \\(surrogate module) \end{tabular} & text & BERT & \cite{109} \\
    ADVD/TYPD & P/F-BefA/AftA & White-box & \begin{tabular}{c} Adversarial Training \\(surrogate module) \end{tabular} & multi & CLIP & \cite{120}\cite{136}\\
    
    TYPD & F-BefA/AftA & White-box & Adversarial Training & image & CLIP & \cite{112} \\
    JBD & P/F-BefA & White-box & Adversarial Training & text & Vicuna & \cite{134} \\
    
    MED & F-BefA & White-box & Adversarial Training & image & ResNet & \cite{83} \\
    \bottomrule
    \end{tabular}
    \begin{tablenotes} \footnotesize \item BDD/ADVD/TYPD/JBD is short for Backdoor/Adversary/Typographic/Jailbreak Defense; MED is short for Model Extraction Defense. \item Stage P/F/I is short for stage Pre-training/Fine-tuning/Inference. Stage BefA/AftA is short for the Before-Attack/After-Attack stage. \end{tablenotes}
\end{table}

In these last few subsections, we reviewed studies that use adversary training as the defense method. A summary of surveyed studies is shown in Table~\ref{tab:Summary_of_Model-Change_Defense}.
From those summaries, we can conclude that 
(1) all Model-Changed defense methods require white-box access to target models, as defenders need the ability to modify target models; and 
(2) these defense methods focus more on the Before-Attack stage than the After-Attack stage.

At the Before-Attack stage,
Some studies \cite{102}\cite{123} investigate the alignment of multimodal models and develop defense strategies targeting both modalities. These works focus on making clean samples in the training stage more closely connected to increase model robustness. 
Some researchers also predict potential future attacks and prepare for them in advance \cite{112}\cite{83}.
Other studies leverage surrogate models as the supervisor to guide the learning process of target models. Some of them \cite{120}\cite{109} introduce newly designed modules to jointly train with target models and continuously provide defense after deployment. In contrast, surrogate models in other works \cite{136}\cite{127}\cite{134} only contribute during the training process. 
These preemptive methods anticipate potential attack methods and incorporate carefully designed optimization targets into the original training process, effectively enhancing the model's robustness

However, due to the rapid development of attack methods, it is unrealistic to defend against all possible attacks before they occur. Instead, defenders can rely on their experience to select and preemptively defend against the most likely attack methods.
Moreover, since defenders introduce new optimization targets into the original training process, this may risk compromising the model’s performance in the original task and increase computational costs

From another perspective, adversary training defense methods can mitigate attack performance post-attack. When an attack occurs, defenders clearly understand the characteristics and forms of the attack and can design targeted defense methods. For example, defenders can interpolate backdoored weights with clean weights \cite{104} or collect known malicious samples \cite{112}\cite{136} to enhance the target model’s robustness.
As it is time-consuming and high-cost for defenders to manually collect all possible malicious input, one good way is to apply the ability of large models to generate malicious input automatically \cite{134}.
This method excels in defense performance since defenders know the malicious samples or attack methods. However, attackers will create new malicious samples, and it's uncertain whether the model can defend against these new attacks. Generating as many malicious samples as possible is a good approach, though more effective defense methods should be explored.

\section{Future Directions}

In this section, we analyze current and potential trends in pre-trained model security and summarize our findings. Furthermore, we identify several unaddressed research directions that could advance further research on the security of pre-trained models.

\paragraph{Attack Transferability to Multimodal Large Pre-trained Models}

Multimodal large pre-trained models support more than one input type. 
Current mainstream models focus on text-modal (GPT-3) and text-image-modal (GPT-4, LLaVA). However, handling only text and image abilities is not enough to make large pre-trained models to be human. Therefore, researchers hope large pre-trained models can understand data from more modalities like audio (X-LLM \cite{chen2023x} and GPT-4o) and video (Video-LLaMA \cite{zhang2023video} and Video-ChatGPT \cite{maaz-etal-2024-video}).
These models add a new branch to process the audio/video modal, both feature extraction ability and the alignment between each modal play an important role in the model performance.

There is no doubt that new modalities will bring new security issues. However, the experience in text-image models can be transferred to multimodal models because of their similar structure, which is the combination of an encoder to process each modal and a LLM to understand each modal and generate final results.
Past works focus on attack transferability between text-modal to text-image-modal by leveraging the connection between multimodal features. \cite{55} summarized two key roles contributing to transferable attacks: (1) modality-consistency features, which represent decision-related characteristics shared among different modalities, and (2) unique characteristics specific to each modality on which the model's decision does not depend. \cite{60} assumed that transferability degradation is partly caused by under-utilization of cross-modal interactions. Particularly, unlike single-modal learning we think these works have transferability due to the alignment of text-image features which is important in building text-image models. This idea can be applied to design new attack strategies for audio-based/video-based large pre-trained models. 

In addition, as most future multimodal large pre-trained models are possible close-source models, the current experience that attacks transferability from open-source model to close-source model can be leveraged to attack multimodal models. There is a consensus that when attackers have white-box access to the model, the designed attack can achieve wonderful attack results. When attackers try to attack closed-source models, there are still various challenges such as poor attack performance and insufficient explainability.
Some works find that generating malicious prompts can attack other close-source LLMs like GPT-4 \cite{1}. \cite{46} optimized adversarial prompts on multiple open-source LLMs to increase the transferability. However, they just give experiment results to show transferability and do not answer the reason why these malicious prompts can be effective in attacking other models. \cite{56} studied the transferability of the same image perturbation across different prompts optimizing an image perturbation using up to 100 text prompts simultaneously.
We find their transferability comes from the specifically designed optimization goal.
As different models have different embedding layers, previous studies injected backdoors into embedding layers or word embedding vectors, limiting the transferability. \cite{80} proposed to directly bind trigger and target anchors into an encoder, bypassing different embedding layers, so that this method can be transferred to different models.

In summary, attack transferability is a difficult but extremely meaningful research direction that can help this field better understand the relationship between attacks and models, thereby better defending and improving the model itself. Furthermore, current works focus on attacking text and image modalities, and studying attack transferability can help us understand future multimodal models focusing on audio and video.

\paragraph{Large Pre-trained Model Hallucination}

Although Large Pre-trained Models have marked a significant breakthrough on various CV and NLP tasks, they exhibit a critical tendency to produce hallucinations, resulting in content that is inconsistent with real-world facts or user inputs \cite{huang2023survey, liu2024survey}. 
Model developer needs to make sure their model won't generate any incorrect/harmful/malicious response, as this may cause financial losses or even loss of user trust. 
This phenomenon poses a great challenge to its practical application and raises concerns about the reliability of large pre-trained models in real-world scenarios, thus attracting increasing attention to detect and mitigate these illusions \cite{farquhar2024detecting, huang2023transformer}.

There are several reasons why model hallucination exists: 
(1) there needs update in the way researchers teach the large pre-trained models to understand user instructions. For example, when users ask GPT-4 ``Which number is bigger? 9.13 or 9.6'', GPT-4 make incorrect judgement that ``9.13 is bigger''. This is because large pre-trained models have some misunderstandings about numbers and do not correctly interpret user instructions, resulting in mistakes. Therefore, future work can focus on this part and study how to correctly encode natural text/numbers/images into features. The difficulty lies in that 26 letters can be combined into different words, and thus into large amounts of sentences with different meanings. However, for large pre-trained models, a sentence is just a combination of 26 letters. To establish a mapping from letters to sentences is a problem worth studying.
(2) Inaccurate results may also be due to large pre-trained models' lack of knowledge in a certain area. Similar to humans, it's hard for humans to answer questions they don't know the answer. To solve this, one way is to collect the newest knowledge in the world and force large pre-trained models to learn this knowledge. However, the continuous development of knowledge also means that models need to be constantly updated. Another way is to leverage retrieval-augmented generation to retrieve relevant information from an external knowledge source, enabling large pre-trained models to answer questions about private and/or previously unseen document collections \cite{RAG1, RAG2}.
(3) large pre-trained models are too powerful to generate answers to any questions based on learned knowledge. However, we don't know if this generated answer is totally correct or partially correct. An example is that large pre-trained models can generate code that runs successfully but does not meet all requirements. One solution is to make models more confident in their response through RLHF \cite{ouyang2022training}. However, this process is time-consuming as large pre-trained models can not replace humans totally to do this work.

\paragraph{Attack Pre-trained Models for Specific Area}

In general CV tasks and NLP tasks, the performance of current mainstream large pre-trained models is close to or even better than humans. This success is due to large pre-trained datasets collected from websites and carefully labeled specific datasets used for fine-tuning. 
Many works extend the general research field to more specific scenarios, such as medicine \cite{waisberg2024large}, psychology \cite{demszky2023using}, etc. These works focus on more professional fields and design large pre-trained models according to actual needs in such fields to liberate the need for traditional human-based work, so it is a broad research direction.

However, various characteristics of different research fields are also exploited by attackers to design different attack methods. For example, in the medical domain, malicious attackers can use large pre-trained models to generate a malicious paper that poisons medical knowledge graphs constructed from millions of real papers, causing consumers of the poisoned knowledge graph will misidentify this promoting drug as relevant to the target disease \cite{yang2024poisoning}. 
There are several differences between area-specific large pre-trained models and general large pre-trained models: (1) The response of area-specific large pre-trained models must be correct, otherwise there will be serious consequences. However, since general large pre-trained models have to learn too much knowledge, they can't be professional in every field. (2) Area-specific large pre-trained models are toolkits to assist work in specific tasks, and human is still required, while general large pre-trained models can deliver most tasks alone and these tasks are relatively easy.
To investigate the attack/defense issues in a specific area,
researchers should first be area experts and find specific issues that are valuable to attack/defense.
Future work can use large pre-trained models to attack specific systems in application fields, discover existing security risks, and build a more robust system.

\paragraph{Machine Unlearning in Large Pre-trained Models}

Since large pre-trained models are trained on billions of datasets, which may contain private or harmful information, model owners need to find methods to make these models forget such information, and machine unlearning is an effective approach. 
The purpose of Machine Unlearning is to make the target model forget some training data \cite{10.1145/3603620}. Its effect has been effectively verified under traditional models. Applying machine unlearning to the field of large pre-trained models can (1) remove harmful responses, (2) delete copyrighted content, and (3) reduce hallucinations \cite{LLMU}.

Large pre-trained models are different from traditional models in terms of dataset and model size, as well as new training methods (such as prompt learning, RHLF, etc.), bringing new challenges to machine unlearning tasks. The first challenge is to accurately summarize harmful content (unlearning targets) defined by humans into a specific standard that can effectively guide subsequent research. Second, the growing size of large pre-trained models and the rise of black-box access to these models present challenges in developing scalable and adaptable machine unlearning techniques for large pre-trained models \cite{liu2024rethinking}.

\section{Conclusion}
Pre-trained models have become a powerful driving force for innovation in various applications. However, pre-trained models expose various security vulnerabilities, which hinder their application in real life. In addition, the threats faced by pre-trained models are very different from those faced by traditional models. This has attracted widespread attention from academia and industry.
In this survey, we provide a comprehensive overview of attacks and corresponding defenses. 
From the perspective of attackers, we consider three attack threat models (No-Change Attacks, Input-Change Attacks, and No-Change Attacks) based on the influence of the target model and three attack scenarios (pre-training stage, fine-tuning stage, and inference stage) based on the life process of pre-trained models.
Additionally, from the perspective of defenders, we consider three defense threat models (No-Change Defenses, Input-Change Defenses, and Model-Change Defenses) based on the defenders' ability to target models and two defense scenarios (Before-Attack stage and After-Attack stage). For each condition, we describe common risks and highlight unique risks from small pre-trained models to large pre-trained models as well as possible defense strategies.
After summarizing current work, we propose several further research directions, such as attack transferability on multimodal models, large pre-trained models' hallucination, attack in specific areas, and machine unlearning in large pre-trained models. We believe this survey can effectively help researchers understand the security threats of pre-trained models and inspire them to conduct in-depth research.

\section{Acknowledgement}


\bibliographystyle{ACM-Reference-Format}
%\bibliographystyle{unsrt}
%\bibliography{sample-base}
\bibliography{main}

% \appendix

\end{document}